\documentclass[acmtog, authorversion]{acmart}
\acmSubmissionID{159}
\pdfoutput=1

\usepackage{enumitem}
\usepackage{subcaption}

\newcommand{\noarxiv}[1]{}

\usepackage{accents}
\newlength{\dhatheight}

\usepackage{mathtools}

\usepackage{xspace}
\makeatletter
\DeclareRobustCommand\onedot{\futurelet\@let@token\@onedot}
\def\@onedot{\ifx\@let@token.\else.\null\fi\xspace}

\def\eg{{e.g}\onedot} 
\def\ie{{i.e}\onedot}

\def\etal{{et al}\onedot}

\makeatother

\usepackage{sistyle}
\SIthousandsep{,}

\def\mF{\mathcal{F}}

\def\mV{\mathcal{V}}

\def\1n{\mathbf{1}_n}
\def\0{\mathbf{0}}
\def\1{\mathbf{1}}

\def\R{{\mathbb R}}

\def\f{{\bf f}}

\def\p{{\bf p}}
\def\q{{\bf q}}

\def\x{{\bf x}}

\newcommand{\us}{$u$s\xspace}
\newcommand{\ms}{$m$s\xspace}

\newcommand{\cm}{$c$m\xspace}
\newcommand{\mm}{$m$m\xspace}
\newcommand{\nm}{$n$m\xspace}
\newcommand{\um}{$u$m\xspace}

\newcommand{\rect}{\mbox{rect}}
\newcommand{\sinc}{\mbox{sinc}} \usepackage{physics}

\usepackage{booktabs} %

\setcitestyle{square}

\usepackage{xr}
\usepackage[ruled]{algorithm2e} %

\SetAlFnt{\small}
\SetAlCapFnt{\small}
\SetAlCapNameFnt{\small}
\SetAlCapHSkip{0pt}
\IncMargin{-\parindent}

\begin{document}
\title[Dense Multifocal]{Towards Multifocal Displays with Dense Focal Stacks}

\author{Jen-Hao Rick Chang}
\orcid{1234-5678-9012-3456}
\affiliation{%
  \institution{Carnegie Mellon University}
  \streetaddress{5000 Forbes Ave}
  \city{Pittsburgh}
  \state{PA}
  \postcode{15213}
  \country{USA}}
\email{rickchang@cmu.edu}
\author{B.\ V.\ K.\ Vijaya Kumar}
\affiliation{%
  \institution{Carnegie Mellon University}
  \city{Pittsburgh}
  \country{USA}
}
\email{kumar@ece.cmu.edu}
\author{Aswin C.\ Sankaranarayanan}
\affiliation{%
	\institution{Carnegie Mellon University}
	\city{Pittsburgh}
	\country{USA}
}
\email{saswin@andrew.cmu.edu}

\renewcommand\shortauthors{Chang et al}

\begin{abstract}
We present a virtual reality display that is capable of generating a dense collection of depth/focal planes. 
This is achieved by driving a focus-tunable lens to sweep a range of focal lengths at a high frequency and, subsequently, tracking the focal length precisely at microsecond time resolutions using an optical module.
Precise tracking of the focal length, coupled with a high-speed display, enables our lab prototype to generate 1600 focal planes per second. 
This enables a novel first-of-its-kind virtual reality multifocal display that is capable of resolving the vergence-accommodation conflict endemic to today's  displays.
\end{abstract}

\setcopyright{acmcopyright}
\acmJournal{TOG}
\acmYear{2018}\acmVolume{37}\acmNumber{6}\acmArticle{198}\acmMonth{11} \acmDOI{10.1145/3272127.3275015}

\begin{CCSXML}
<ccs2012>
<concept>
<concept_id>10010147.10010371.10010387.10010866</concept_id>
<concept_desc>Computing methodologies~Virtual reality</concept_desc>
<concept_significance>500</concept_significance>
</concept>
</ccs2012>
\end{CCSXML}

\ccsdesc[500]{Computing methodologies~Virtual reality}

\keywords{focus-tunable lenses, multifocal displays, focus stacks}

\begin{teaserfigure}
   \includegraphics[width=\linewidth]{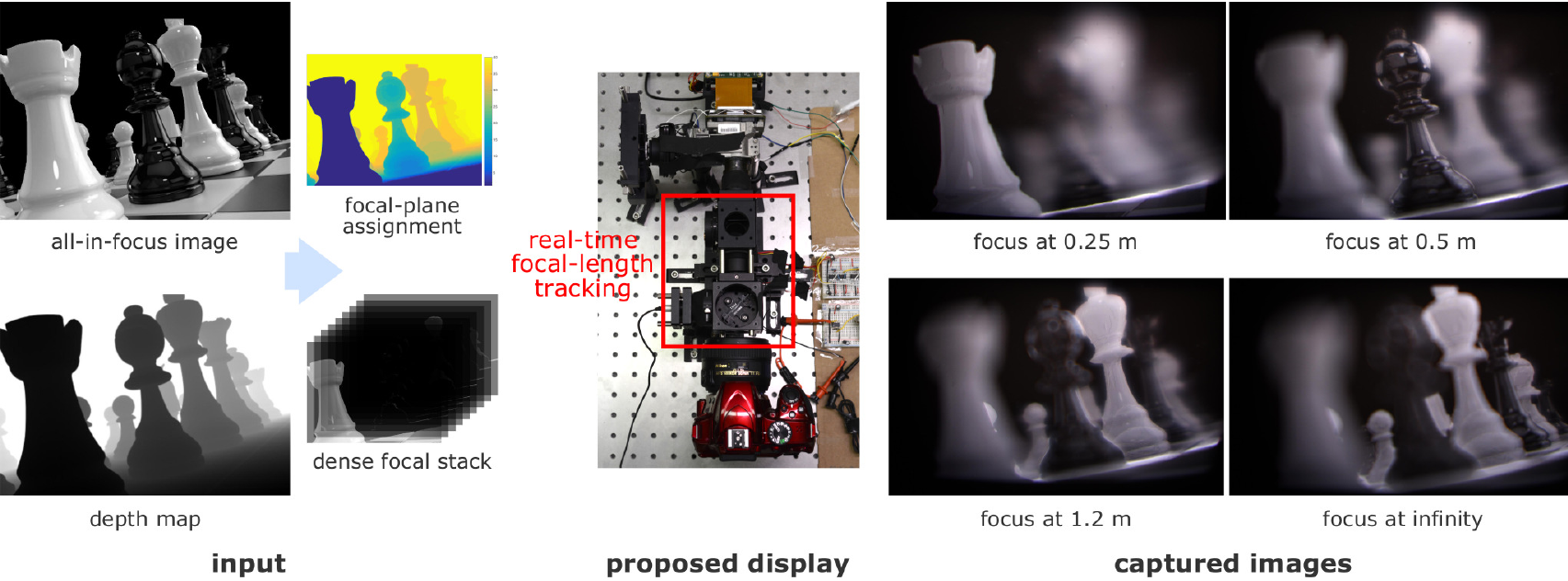}
	\vspace{-7mm}
	\caption{
	Producing strong focusing cues for the human eye requires rendering scenes with dense focal stacks. 
	This would require a virtual reality display that can produce thousands of focal planes per second.
	We achieve this capability by exciting a focus-tunable lens with a high-frequency input and  tracking its focal length at microsecond time resolution.
	Using a lab prototype, we demonstrate that the high-speed tracking of the focal length, coupled with a high-speed display, can render a very dense set of focal stacks. 
	Our system is capable of generating 1600 focal planes per second, which we use to render 40 focal planes per frame at 40 frames per second. 
	Shown are images captured with a 50mm $f/2.8$ lens focused at different depths away from the tunable lens.
}
\end{teaserfigure}

\maketitle

\section{Introduction} \label{sec:intro}

The human eye automatically changes the focus of its lens to provide sharp, in-focus images of objects at different depths.
While convenient in the real world, for virtual or augmented reality (VR/AR) applications, this focusing capability of the eye often causes a problem that is called the vergence-accommodation conflict (VAC)~\cite{kramida2016resolving,hua2017enabling}.
Vergence refers to the simultaneous movement of the two eyes so that a scene point comes into the center of the field of view, and accommodation refers to the changing of the focus of the ocular lenses to bring the object into focus.
In the real world, these two cues act in synchrony.
However, most commercial VR/AR displays render scenes by only satisfying the vergence cue, \ie, they manipulate the disparity of the images shown to each eye.
But given that the display is at a fixed distance from the eyes, the corresponding accommodation cues are invariably incorrect, leading to a conflict between vergence and accommodation that can cause
discomfort, fatigue, and distorted 3D perception,  especially after long durations of usage \cite{hoffman2008vergence,watt2005focus,vishwanath2010retinal,zannoli2016blur}.
While many approaches have been proposed to mitigate the VAC, it remains one of the  important challenges for VR and AR displays.

In this paper, we provide the design for a VR display that is capable of addressing the VAC by displaying content on  a dense collection of depth or focal planes.
The proposed display falls under the category of multifocal displays, \ie, displays that generate content at different focal planes %
using a focus-tunable lens %
\cite{liu2008optical,liu2009time,love2009high,llull2015design,johnson2016dynamic,konrad2016novel}.
This change in focal length can be implemented in one of many ways; for example, by changing the curvature of a liquid lens~\cite{optotune,varioptic}, the state of a liquid-crystal lens~\cite{Jamali18,jamali2018design}, the polarization of a waveplate lens~\cite{tabiryan2015thin}, or the relative orientation between two carefully designed phase plates~\cite{bernet2008adjustable}.
The key distinguishing factor is that the proposed device displays a stack of focal planes that are an order of magnitude greater in number as compared to prior work, without any loss in the frame rate of the display.
Specifically, our prototype system is capable of displaying 1600 focal planes per second, which can be used to display scenes with 40 focal planes per frame at 40 frames per second.
As a consequence, we are able to render virtual worlds at a realism that is hard to achieve with current multifocal display designs.

To understand how our system can display thousands of focal planes per second, 
it is worth pointing out that the key 
 factor that limits the depth resolution of a multifocal display is the operational speed of its focus-tunable lens.
Focus-tunable liquid lenses change their focal length based on an input driving voltage; they typically require  around $5$~\ms to settle onto a particular focal length.
Hence, in order to wait for the lens to settle so that the displayed image is rendered at the desired depth, we can output at most  $200$ focal planes per second. 
For a display operating with 30-60 frames per second (fps), this would imply anywhere between three and six focal planes per frame, which is woefully inadequate.

The proposed display relies on the observation that, while focus-tunable lenses have long settling times, their frequency response is rather broad and has a cut-off upwards of $1000$ Hz~\cite*{optotune}.
This suggests that we can drive the lens with excitations that are radically different from  a simple step edge (\ie, a change in voltage).
For example, we could make the lens sweep through its entire gamut of focal lengths at a high frequency simply by exciting it with a sinusoid or a triangular voltage of the desired frequency.
If we can subsequently track the focal length of the lens in real-time, we can accurately display focal planes at any depth without waiting for the lens to settle.
In other words, by driving the focus-tunable lens to periodically sweep the desired range of focal lengths and tracking the focal length at high-speed and in real-time, we can display numerous focal planes.
\subsection{Contributions} 
This paper proposes the design  of a novel multifocal display  that produces three-dimensional scenes by displaying dense focal stacks.  In this context, we make the following contributions:
\begin{itemize}[leftmargin=*]
	\item \textit{High-speed focal-length tracking.} The core contribution of this paper is a system for real-time tracking of the focal length of a focus-tunable lens at microsecond-scale resolutions. We achieve this by measuring the deflection of a laser incident on the lens.
	\item \textit{Design space analysis.} Displaying a dense set of focal planes is also necessary for mitigating the loss of spatial resolution due to the defocus blur caused by the ocular lens. To show this, we  analytically derive the  spatial resolution of the image formed on the retina when there is a mismatch between the focus of the eye and the depth at which the content is virtually rendered. This analysis justifies the need for AR/VR displays capable of a high focal-plane density.

\item \textit{Prototype.} Finally, we build a proof-of-concept prototype that is able to produce $40$ $8$-bit focal planes per frame with $40$ fps. This corresponds to $1600$ focal planes per second --- a capability that is an order of magnitude greater than competing approaches.
\end{itemize}

\def\figurehieght{\linewidth}
\begin{figure*}[!thh]
\centering
\begin{subfigure}[t]{0.245\linewidth}
\centering
\includegraphics[height=\figurehieght]{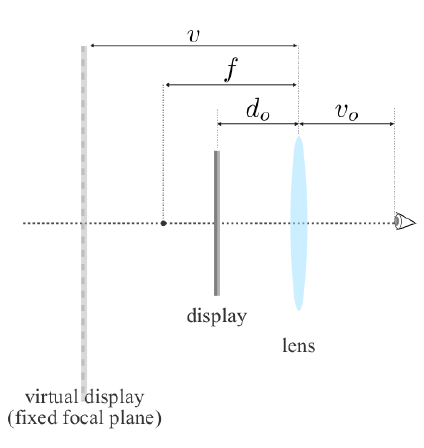}
\caption{\footnotesize Typical VR display}
\end{subfigure}	
~
\begin{subfigure}[t]{0.245\linewidth}
	\centering
	\includegraphics[height=\figurehieght]{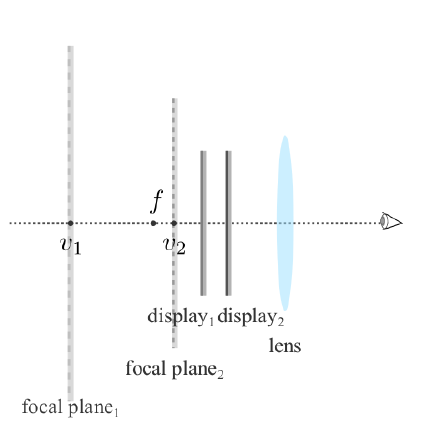}
	\caption{\footnotesize Multifocal with fixed displays}
\end{subfigure}	
~
\begin{subfigure}[t]{0.245\linewidth}
	\centering
	\includegraphics[height=\figurehieght]{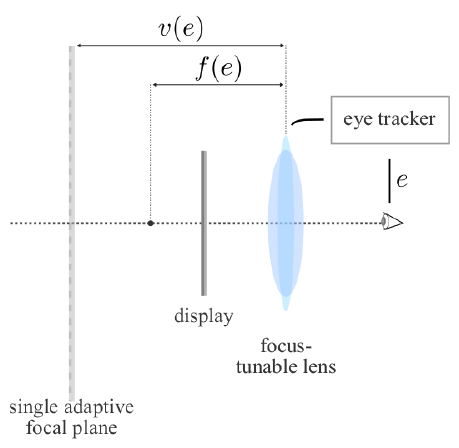}
	\caption{\footnotesize Varifocal with a focus-tunable lens}
\end{subfigure}	
~
\begin{subfigure}[t]{0.245\linewidth}
	\centering
	\includegraphics[height=\figurehieght]{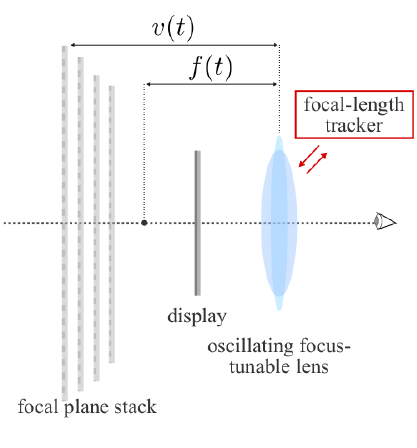}
	\caption{\footnotesize Proposed display}
\end{subfigure}		
\caption{Typical VR displays have a fixed display with a fixed focal-length lens and thereby can output one focal plane at a fixed depth.  Multifocal displays can produce multiple focal planes within a frame, using either multiple displays (shown above) or liquid lenses.  Varifocal displays generate a single but adaptive focal plane using an eye tracker.  The proposed display  outputs dense focal plane stacks by tracking the focal-length of an oscillating focus-tunable lens. The depths of the focal planes are independent to the viewer, and thereby eye trackers are optional.}
\label{figure: displays}
\end{figure*}

\subsection{Limitations} 
In addition to limitations endemic to multifocal displays, the proposed approach has the following limitations:
\begin{itemize}[leftmargin=*]
\item  \textit{Need for additional optics.}  The proposed focal-length tracking device requires additional optics  that increase its bulk.
\item \textit{Peak brightness.}  Displaying a large number of focal planes per frame leads to a commensurate decrease in peak brightness of the display since each depth plane  is illuminated for a smaller fraction of time. This is largely not a concern for VR displays, and can potentially be alleviated with techniques that redistribute light~\cite{damberg2016high}.

\item \textit{Limitations of our prototype.}
Our current proof-of-concept prototype uses a digital micromirror display (DMD) and, as a consequence, has low energy efficiency.
The problem can be easily solved by switching to energy-efficient displays, like OLED, or laser-scanning projectors or displays that redistribute light to achieve higher peak brightness and contrast.

\end{itemize}

\section{Related work}

A typical VR display is composed of a convex eyepiece and a display unit.  
As shown in Figure~\ref{figure: displays}a, the display is placed within the focal length of the convex lens in order to create a magnified virtual image. 
The distance $v > 0$ of the virtual image can be calculated by the thin lens formula: 
\begin{equation}
\frac{1}{d_o} + \frac{1}{-v} = \frac{1}{f},
\label{eq: thin lens}
\end{equation}
where $d_o$ is the distance between the display and the lens, and $f$ is the focal length.
We can see that $\frac{1}{v}$ is an affine function of the optical power ($1/f$) of the lens and the term $1/d_o$.
By choosing $d_o$ and $f$, the designer can put the virtual image of the display at the desired depth.
However, for many  applications, most scenes need to be  rendered across a wide range of depths.
Due to the fixed focal plane, these displays do not provide natural accommodation cues.

\subsection{Accommodation-Supporting Displays}

There have been many designs proposed to provide accommodation support.
We concentrate on techniques most relevant to the proposed method, deferring a detailed description to \cite{kramida2016resolving} and \cite{hua2017enabling}; in particular, see Table~1 of \cite{matsuda2017focal}.

\subsubsection{Multifocal and Varifocal Displays}

Multifocal and varifocal displays control the depths of the focal planes by dynamically adjusting $f$ or $d_o$ in \eqref{eq: thin lens}.
Multifocal displays aim to produce multiple focal planes at different depths for each frame (Figure~\ref{figure: displays}b), whereas varifocal displays support only one focal plane per frame whose depth is dynamically adjusted based on the gaze of the user's eyes (Figure~\ref{figure: displays}c). 
Multifocal and varifocal displays can be designed in many ways, including the use of multiple (transparent) displays placed at different depths~\cite{rolland1999dynamic,akeley2004stereo,love2009high}, a translation stage to physically move a display or optics~\cite{shiwa1996proposal,sugihara1998system,Aksit}, deformable mirrors~\cite{hu2014high}, as well as a focus-tunable lens to optically reposition a fixed display~\cite{liu2008optical,padmanaban2017optimizing,johnson2016dynamic,konrad2016novel,lee2018tomo}.
Varifocal focal displays show a single focal plane at any point in time, but they require precise eye/gaze-tracking at low latency.
Multifocal displays, on the other hand, have largely been limited to displaying a few focal planes per frame due to the limited switching speed of translation stages and focus-tunable lenses.
Concurrent to our work, Lee~\etal~\citeNN{lee2018tomo} propose a multifocal display that can also display dense focal stacks with a focus-tunable lens.
However, their method can only display any given pixel at a single depth. 
This prohibits the use of rendering techniques~\cite{akeley2004stereo,narain2015optimal,mercier2017multifocal} that require a pixel to be potentially displayed at many depths with different contents.

\subsubsection{Light Field Displays}

While multifocal and varifocal displays produce a collection of focal planes, light field displays aim to synthesize  the light field of a 3D scene.
Lanman and Luebke~\citeNN{lanman2013near} introduce angular information by replacing the eyepiece with a microlens array; Huang~\etal~\citeNN{huang2015light} utilize multiple spatial light modulators to modulate the intensity of light rays.
While these displays fully support accommodation cues and produce natural defocus blur and parallax, they usually suffer from poor spatial resolution due to the space-angle resolution trade-off.

\subsubsection{Other Types of Virtual Reality Displays}
Other types of VR/AR displays have been proposed to solve the VAC. %
Matsuda~\etal~\citeNN{matsuda2017focal} use a phase-only spatial light modulator to create spatially-varying lensing based on the virtual content and the gaze of the user.
Maimone~\etal~\citeNN{maimone2017holographic} utilize a phase-only spatial light modulator to create a 3D scene using holography.
Similar to our work, Konrad~\etal \citeNN{konrad2017accommodation}  operate a focus-tunable lens in an oscillatory mode.
Here, they  use the focus-tunable lens to create a depth-invariant blur by using a concept proposed for extended depth of field imaging~\cite{miau2013focal}.
Intuitively, since the content is displayed at all focal planes, the VAC is significantly resolved.
However, there is a loss of spatial resolution due to the intentionally introduced defocus blur.

\subsection{Depth-Filtering Methods}
\label{sec: depth filtering}

When virtual scenes are rendered with few focal planes, there are associated aliasing artifacts as well as a reduction of spatial resolution on content that is to be rendered in between  focal planes.
\citeN{akeley2004stereo} show that such artifacts can be alleviated using linear depth filtering,  a method that is known to be quite effective~\cite{mackenzie2010accommodation,ravikumar2011creating}.
However, linear depth filtering produces artifacts near object boundaries due to the inability of multifocal displays to occlude light.
To produce proper occlusion cues with multifocal displays, \citeN{narain2015optimal} propose a method that jointly optimizes  the contents shown on all focal planes. 
By modeling the defocus blur of focal planes when an eye is focused at certain depths, they formulate a non-negative least-square problem that minimizes the mean-squared error between perceived images and target images at multiple depths.
While this algorithm demonstrates promising results, the computational costs of the optimization are often too high for real-time applications.
\citeN{mercier2017multifocal} simplify the forward model of \citeN{narain2015optimal} and significantly improve the speed to solve the optimization problem.
These filtering approaches are largely complementary to the proposed work, in that,  they can be incorporated into the dense focal stacks produced by our proposed display.

\section{How Many Focal Planes Do We need?} \label{sec: how many}

\begin{figure*}[!thh]
	\centering
	\includegraphics[width=\linewidth]{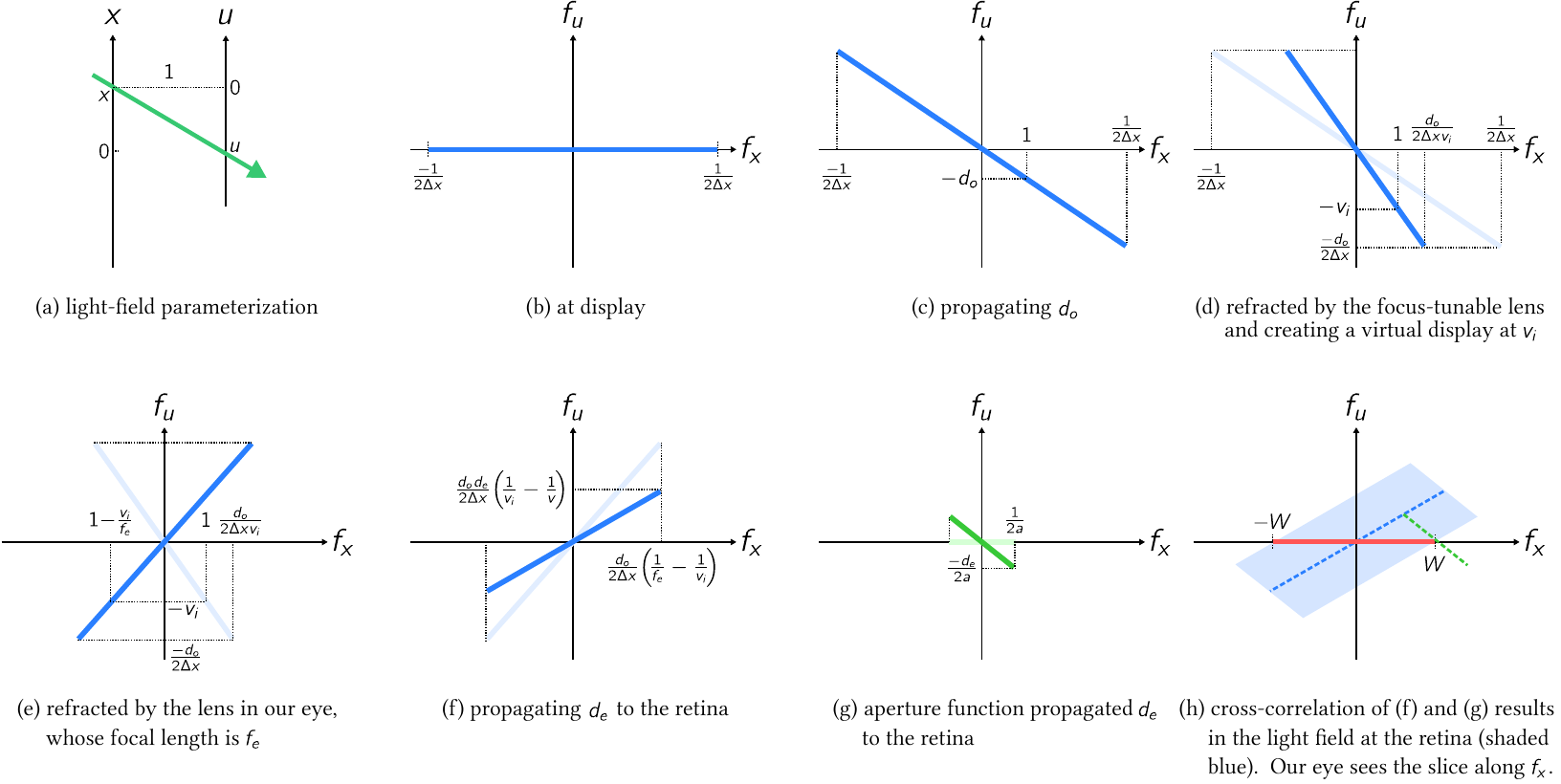}
	\caption{Fourier transform of the 2-dimensional light field at each stage of a multifocal display.  The display is assumed to be isotropic and has pixels of pitch $\Delta x$.  (a) Each light ray in the light field  is characterized by its intercepts with two parallel axes, $x$ and $u$, which are separated by $1$ unit, and the origin of the $u$-axis is relative to each individual value of $x$. (b) With no angular resolution, the light field spectrum emitted by the display is a flat line on $f_x$.  We focus only on the central part ($|f_x| \le \frac{1}{2\Delta x}$).  (c) The light field propagates $d_o$ to the tunable lens, causing the spectrum to shear along $f_u$.  (d) Refraction due to the lens corresponds to shearing along $f_x$, forming a line segment of slope $-v_i$, where $v_i$ is the depth of the focal plane.  (e,f) Refraction by the lens in our eye  and propagation $d_e$ to the retina without considering the finite aperture of the pupil.  (g) The spectrum of the pupil function propagates $d_e$ to the retina.  (h) The light field spectrum on the retina with a finite aperture is the 2-dimensional cross-correlation between (f) and (g).  According to Fourier slice theorem, the spectrum of the perceived image is the slice along $f_x$, shown as the red line.  The diameter of the pupil and the slope of (f), which is determined by the focus of the eye and the virtual depth $v_i$, determine the spatial bandwidth, $W$, of the perceived image.  }
	\label{figure: light field}
\end{figure*}
A key factor underlying the design of multifocal displays is the number of  focal planes required to support a target accommodation range.
In order to be indistinguishable from the real world, a virtual world should enable human eyes to  accommodate freely on arbitrary depths.
In addition, the virtual world should have high spatial resolution anywhere within the target accommodation range.
Simultaneously satisfying these two criteria for a large accommodation range is very challenging, since it requires generating light fields of high spatial and angular resolution.
In the following, we will show that displaying a dense focal stack is a promising step toward the ultimate goal of generating  virtual worlds that can handle the accommodation cues of the human eye.

To understand the capability of a multifocal display, we can analyze its generated light field in the frequency domain.
Our analysis, following the derivation in \citeN{wetzstein2011layered} and \citeN{narain2015optimal}, provides an upper-bound on the performance of a multifocal display, regardless of the depth filtering algorithm applied.  
It is also similar to that of~\citeN{sun2017perceptually} with the key difference that we focus on the minimum number of focal planes required to retain spatial resolution within an accommodation range,  as opposed to efficient rendering of foveated light fields.
\subsection{Light-Field Parameterization and Assumptions}
For simplicity, our analysis considers a flatland with two-dimensional light fields.
In the flatland, the direction of a light ray is parameterized by its intercepts with two parallel axes, $x$ and $u$, which are separated by $1$ unit, and the origin of the $u$-axis is relative to each individual value of $x$ such that $u$ measures the tangent angle of a ray passing through $x$, as shown in Figure~\ref{figure: light field}a.
We model the human eye with a camera composed of a finite-aperture lens and a sensor plane $d_e$ away from the lens, following the assumptions made in~\citeN{mercier2017multifocal} and~\citeN{sun2017perceptually}.
We assume that the pupil of the eye is located at the center of the focus-tunable lens and is smaller than the aperture of the tunable lens. 
We assume that the display and the sensor emits and receives light isotropically.  
In other words, each pixel on the display uniformly emits light rays toward every direction and vice versa for the sensor.
We also assume small-angle (paraxial) scenarios, since the distance $d_o$ and the focal length of the tunable lens (or essentially, the depths of focal planes) are large compared to the diameter of the pupil.
This assumption simplifies our analysis by allowing us to consider each pixel in isolation.
\subsection{Light Field Generated by the Display}
Since the display is assumed to emit light isotropically in angle, the light field created by a display pixel can be modeled as $\ell_d(x,u) = I\ \delta(x) \ast \rect\left(\frac{x}{\Delta x}\right)$, where $I$ is the radiance emitted by the pixel, $\ast$ represents two-dimensional convolution, and $\Delta x$ is the pitch of the display pixel.
The Fourier transform of $\ell_d(x,u)$ is $L_d(f_x, f_u) = \frac{I}{\Delta x} \sinc(\Delta x f_x)$, which lies on the $f_x$ axis, as shown in Figure~\ref{figure: light field}b. %
We only plot the central lobe of $\sinc(\Delta x f_x)$ corresponding to  $|f_x| \le \frac{1}{2 \Delta x}$, since this is sufficient for calculation of the half-maximum bandwidth of retinal images.
In the following, we omit the constant $\frac{I}{\Delta x}$ for brevity.

\subsection{Propagation from Display to Retina}
Let us decompose the optical path from the display to the retina (sensor) and examine its effects in the frequency domain.
After leaving the display, the light field propagates a distance $d_o$, gets refracted by the tunable lens, and by the lens of the eye where it is partially blocked by the pupil, whose diameter is $a$, and propagates a distance $d_e$ to the retina where it finally gets integrated across angle.
Propagation and refraction shears the spectrum of the light field along $f_u$ and $f_x$, respectively, as shown in Figure~\ref{figure: light field}(c,d,e).
Before entering the pupil, the focal plane at depth $v_i$ forms a segment of slope $-v_i$ within $|f_x| \le \frac{d_o}{2 v_i \Delta x}$, where $\frac{d_o}{v_i}$ is due to the magnification of the lens.
For brevity, we show only the final (and most important) step and defer the full derivation to the appendix.

Suppose the eye focuses at depth $v=f_e d_e/(d_e-f_e)$, and the focus-tunable lens configuration creates a focal plane at $v_i$.
The Fourier transform of the light field reaching the retina is 
\begin{equation}
L_e(f_x, f_u) = L^{(v_i)}(f_x, f_u) \otimes A^{(d_e)}(f_x, f_u),
\end{equation}
where $\otimes$ represents two-dimensional cross correlation, $L^{(v_i)}$ is the Fourier transform of the light field from the focal plane at $v_i$ reaching the retina without aperture (Figure~\ref{figure: light field}f), and $A^{(d_e)}$ is the Fourier transform of the aperture function propagated to the retina (Figure~\ref{figure: light field}g). %
Depending on the virtual depth $v_i$, the cross correlation creates different extent of blur on the spectrum (Figure~\ref{figure: light field}h).
Finally, the Fourier transform of the image that is seen by the eye is simply the slice along $f_x$ on $L_e$.

When the eye focuses at the focal plane ($v = v_i$), the spectrum lies entirely on $f_x$ and the cross correlation with $A^{(d_e)}$ has no effect on the spectrum along $f_x$.  
The resulted retinal image has maximum spatial resolution $\frac{d_o}{2 d_e \Delta x}$, which is independent of the depth of the focal plane $v_i$.

When the eye is not focused on the virtual depth plane, \ie, $v_i \neq v$,  the cross correlation results in a segment of width \[ W = \frac{1}{2 a d_e}  \left( \left| \frac{1}{v} - \frac{1}{v_i} \right| \right)^{-1}\]
 on the $f_x$-axis (Figure~\ref{figure: light field}h).
Note that $|L_e(\pm W, 0)| = \sinc(0.5) \times \sinc(0.5) \approx 0.4$, and thereby the half-maximum bandwidth of the spatial frequency of the perceived image is upper-bounded by~$W$.

\subsection{Spatial Resolution of Retinal Images}
We can now characterize the spatial resolution of a multifocal display.  
Suppose the eye can accommodate freely on any depth $v$ within a target accommodation range, $[v_a, v_b]$.
Let $\mV = \{v_1 = v_a, v_2, \dots, v_n = v_b\}$ be the set of depth of the focal planes created by the multifocal display. 
When the eye focuses at $v$, the image formed on its retina has spatial resolution of
\begin{equation}
F_s(v) = \min \left\{ \frac{d_o}{2 d_e \Delta x},  \ \max_{v_i \in \mV}   \left( 2 a d_e \left| \frac{1}{v} - \frac{1}{v_i} \right| \right)^{-1}  \right\},
\label{eq: spatial res}
\end{equation} 
where the first term characterizes the inherent spatial resolution of the display unit, and the second term characterizes spatial resolution limited by accommodation, i.e. potential mismatch between the focus plane of the eye and the display.
This bound on spatial resolution is a physical constraint caused by the finite display pixel pitch and the limiting aperture (\ie, the pupil) --- even if the retina had infinitely-high spatial sampling rate. 
\textit{Any post-processing methods including linear depth filtering, optimization-based filtering, and nonlinear deconvolution cannot surpass this limitation.}

\subsection{Minimum Number of Focal Planes Needed}
As can be seen in \eqref{eq: spatial res}, the maximum spacing between any two focal planes in diopter determines $\min_{v \in [v_a,v_b]} F_s(v)$, the lowest perceived spatial resolution within the accommodation range.
If we desire a multifocal display with spatial resolution across the accommodation range to be at least $F$, $F \le \frac{d_o}{2 d_e \Delta x}$, the best we can do with $n$ focal planes is to have a constant inter-focal separation in diopter.
This results in an inequality that 
\begin{equation}
\left( \frac{2 a d_e}{2n} \left( \frac{1}{v_a} - \frac{1}{v_b} \right) \right)^{-1}  \ge F,
\end{equation}
or equivalently
\begin{equation}
n \ge a d_e \left(\frac{1}{v_a} - \frac{1}{v_b} \right) F.
\label{eq: num planes}
\end{equation}
Thereby, increasing the number of focal planes $n$ (and distributing them uniformly in diopter) is required for multifocal displays to support higher spatial resolution and wider accommodation range.

\subsection{Relationship to Prior Work.} 
There are many prior works studying the minimum focal-plane spacing of multifocal displays.
Rolland~\etal~\citeyearpar{rolland1999dynamic} compute the depth-of-focus based on typical acuity of human eyes ($30$ cycles per degree) and pupil diameter~($4$~\mm) and conclude that $28$ focal planes equally spaced by $\frac{1}{7}$ diopter are required to accommodate from $25$~\cm to~$\infty$.
Both theirs and our analyses share the same underlying principle --- maintaining the minimum resolution seen by the eye within the accommodation range, and thereby provide the same required focal planes.
By taking $a = 4$~\mm, $d_e F = 30 \times \frac{180}{\pi}$, $v_a = 25$~\cm, and $v_b = \infty$, we have $n \ge 27.5$, which concurs with their result.  
MacKenzie~\etal~\citeyearpar{mackenzie2010accommodation,mackenzie2012vergence} measure accommodation responses of human eyes during usage of multifocal displays with different plane-separation configurations under linear depth filtering~\cite{akeley2004stereo}.
Their results suggest that focal-plane separations as wide as $1$ diopter can drive accommodation with insignificant deviation from the natural accommodation.
However, it is also reported that smaller plane-separations provide more natural accommodation and higher retinal contrast --- features that are desirable in any VR/AR display.
By enabling dense focal stacks of focal-plane separation as small as $0.1$ diopter, our prototype can simultaneously provide proper accommodation cues and display high-resolution  images onto the retina.

\subsection{Maximum Number of Focal Planes Needed}
At the other extreme, if we have a sufficient number of focal planes, the limiting factor becomes the pixel pitch of the display unit.
In this scenario, 
for a focal plane at virtual depth $v_i$, the retinal image of an eye focuses on $v$ will have maximal spatial resolution $\frac{d_o}{2 d_e \Delta x}$ if 
\[ 
\left| \frac{1}{v} - \frac{1}{v_i} \right| \le \frac{ \Delta x}{a d_o}. 
\]
In other words, the depth-of-field of a focal plane --- defined as the depth range that under focus provides the maximum resolution ---  is $\frac{2 \Delta x}{a d_o}$ diopters.
Since the maximum accommodation range of the multifocal display with a convex tunable lens is $\frac{1}{d_o}$ diopter, we need at least $\frac{a}{2 \Delta x}$ focal planes to achieve the maximum spatial resolution of the multifocal display across the maximum supported depth range, or $\frac{D_o a d_o}{2 \Delta x}$ focal planes for a depth range of $D_o$.
For example, our prototype has $\Delta x = 13.6$ \um, $d_o = 7$ \cm, and pupil diameter $a = 4$ \mm, it would require $147$ focal planes for the maximum possible depth range of $d_o = 7$ \cm to infinity or $D_o = 14.3$ diopters to reach the resolution upper-bound.
For a shorter working range of 25 \cm to infinity, or 4 diopters, it would require 41 focal planes.

\section{Generating Dense Focal Stacks}

We now have a clear goal --- designing a multifocal display supporting a very dense focal stack, which enables display high-resolution images across a wide accommodation range.
The key bottleneck for building multifocal displays with dense focal stacks is the settling time of the focus-tunable lens.
The concept described in this section outlines an approach to mitigate this bottleneck and provides a design template for displaying dense focal stacks.

\subsection{Focal-Length Tracking}
The centerpiece of our proposed work is the idea that we do not have to wait for the focus-tunable lens to settle at a particular focal length.
Instead, if we constantly drive the lens so that it sweeps across a range of focal lengths, and subsequently track the focal length in real time,  we can display the corresponding focal plane without waiting for the focus-tunable lens to settle.
This enables us to display as many focal planes as we want, as long as the display supports the required frame rate.

While the optical power of focus-tunable lenses is controlled by an input voltage or current, simply measuring these values only provides inaccurate and biased estimates of the focal length.
This is due to the time-varying transfer functions of tunable lenses, which are known to be sensitive to operating temperature and irregular motor delays. 
Instead, we propose to estimate the focal length by probing the tunable lens optically.
This enables robust estimations that are invulnerable to the unexpected factors.

\begin{figure}[t]
	\centering
	\begin{subfigure}[t]{0.6\linewidth}
		\centering
		\includegraphics[width=\linewidth]{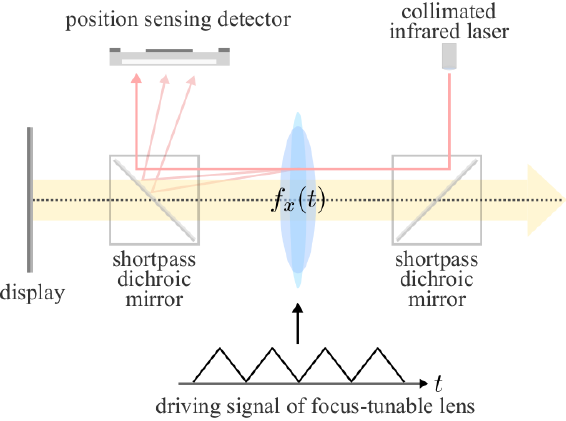}
		\caption{\footnotesize Schematic of focal-length tracking }
	\end{subfigure}	
	~
	\begin{subfigure}[t]{0.39\linewidth}
		\centering
		\includegraphics[width=\linewidth]{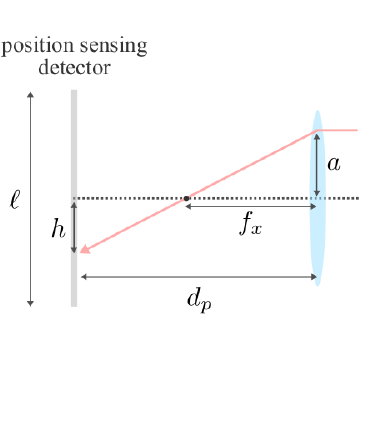}
		\caption{\footnotesize Optical layout}
	\end{subfigure}	
	\vspace{-3mm}
	\caption{(a) The focal-length tracking system is composed of two shortpass dichroic mirrors and a position sensing detector. The dichroic mirror allows visible light to pass through but reflects the infrared light ray emitted from the collimated laser.  (b) The position of the laser spot on the position sensing detector is an affine function of the optical power of the lens.  }
	\label{fig: track}
\end{figure}

In order to measure the focal length, we send a collimated infrared laser beam through the edge of the focus-tunable lens.
Since the direction of the outgoing beam depends on the focal length, the laser beam changes direction as the focal length changes. 
There are many approaches to measure this change in direction, including using a one-dimensional pixel array or an encoder system.
In our prototype, we use a one-dimensional position sensing detector (PSD) to enable fast and accurate measurement of the location.
The schematic is shown in Figure~\ref{fig: track}a.

The focal length of the laser is estimated as follows.
We first align the laser so that it is parallel to the optical axis of the focus-tunable lens.
After deflection by the lens, the beam is incident on a spot on the PSD whose position, 
as shown in Figure~\ref{fig: track}b, is given as
\begin{equation}
 h = a\left(\frac{d_p}{f_x} - 1\right),
 \label{eq: psd_h}
 \end{equation} 
where $f_x$ is the focal length of the lens, $d_p$ is the distance measured along the optical axis between the lens and the PSD, and $h$ is the distance between the optical center of the lens and the spot the laser is incident on.
Note that the displacement $h$ is an affine function of the optical power of the focus-tunable lens.

We next discuss how the location of the spot is estimated from the PSD outputs.
A PSD is composed of a photodiode and a resistor distributed throughout the active area.
The photodiode has two connectors at its anode and a common cathode.
Suppose the total length of the active area of the PSD is $\ell$.
When a light ray reaches a point at $h$ on the PSD,  the generated photocurrent will flow from each anode connector to the cathode with amount inversely proportional to the resistance in between.
Since resistance is proportional to length, we have the ratio of the currents in the anode and cathode as
\begin{equation}
\frac{i_1}{i_2} = \frac{R_2}{R_1} = \frac{\frac{\ell}{2} - h}{\frac{\ell}{2} + h}, \mbox{  or } \ h = \frac{\ell}{2} \frac{i_2 - i_1}{i_2 + i_1}.
\label{eq: trackeq}
\end{equation}
Combining \eqref{eq: trackeq} and~\eqref{eq: psd_h}, we have 
\begin{equation}
\frac{1}{f_x} = \frac{\ell}{2 a d_p} r  + \frac{1}{d_p}, \mbox{  where } r = \frac{i_2 - i_1}{i_2 + i_1} .
\label{eq: 1_fx}
\end{equation}
As can be seen, the optical power of the tunable lens $\frac{1}{f_x}$ is an affine function of $r$. 
With simple calibration (to get the two coefficients),  we can easily estimate the value.

\subsection{The Need for Fast Displays}

In order to display multiple focal planes within one frame, we also require a display that has a frame rate greater than or equal to the focal-plane display rate.
To achieve this, we use a digital micromirror device (DMD)-based projector as our display. 
Commercially available DMDs can easily achieve upwards of $20,000$ bitplanes per second.
Following the design in~\cite{gray}, we modulate the intensity of the projector's light source to display 8-bit images; this enables us to display each focal plane with 8-bits of intensity and generate  as many as  $20,000/8 \approx 2,500$ focal planes per second.

\subsection{Design Criteria and Analysis}
We now analyze the system in terms of various desiderata and the system configurations required to achieve them.

\subsubsection{Achieving a  Full Accommodation Range} 
A first requirement is that the system be capable of supporting the full accommodation range of typical human eyes, \ie, generate focal planes from $25$ \cm to infinity.
Suppose the optical power of the focus-tunable lens ranges from $D_1 = \frac{1}{f_1}$ to $D_2 = \frac{1}{f_2}$ diopter.
From \eqref{eq: thin lens}, we have 
\begin{equation}
\frac{1}{-v(t)} =  \frac{1}{f_x(t)} - \frac{1}{d_o} = -\left( \frac{1}{d_o} - D_x(t) \right), 
\label{eq: Dx}
\end{equation}
where $d_o$ is the distance between the display unit and the tunable lens, $v(t)$ is the distance of the virtual image of the display unit from the lens, $f_x(t) \in [f_2, f_1]$ is the focal length of the lens at time $t$, and $D_x(t) = \frac{1}{f_x(t)}$ is the optical power of the lens in diopter.
Since we want $v(t)$ to range from $25$ cm to infinity, $1/v(t)$ ranges from $4 \textrm{ m}^{-1}$ to $0 \textrm{ m}^{-1}$.
Thereby, we need 
\[
4 - D_1 \le \frac{1}{d_o} \le  D_2.
\]
An immediate implication of this is that $D_2-D_1 \geq 4$, \ie, to support the full accommodation range of a human eye, we need a focus-tunable lens whose optical power spans at least $4$ diopters.
We have more choice over the actual range of focal lengths taken by the lens.
A simple choice is to set $1/f_2 = D_2 = 1/d_o$; this ensures that we can render focal planes at infinity;  subsequently, we choose $f_1$ sufficiently large to cover $4$ diopters.
By choosing a small value of $f_2$, we can have a small $d_o$ and thereby achieve a compact display.

\subsubsection{Field-of-View}
The proposed display shares the same field-of-view and eye box characteristics with other multifocal displays.
The field-of-view will be maximized when the eye is located right near the lens.
This will results in a field-of-view of $2 \atan(\frac{H}{2 d_o})$, where $H$ is the height (or width) of the physical display (or its magnification image via lensing). 
When the eye is further away from the lens, the numerical aperture will limit the extent of the field-of-view.
Since the apertures of most tunable lenses are small (around 1 \cm in diameter), we would prefer to put the eye as close as the lens as possible.  
This can be achieved by embedding the dichroic mirror (the right one in Figure~\ref{fig: track}a) onto the rim of the lens.
For our prototype that will be described in Section~\ref{sec: prototype}, we use a 4$f$ system to relay the eye to the aperture of the focus-tunable lens.  
Our choice of the 4$f$ system enables a $45$-degree field-of-view, limited by the numerical aperture of the lens in the 4$f$ system. 

There are alternate implementations of focus tunable lenses that have the potential for providing larger apertures and hence, displays with larger field of views.
Bernet and Ritsh-Marte~\citeNN{bernet2008adjustable} design two phase plates that produce the phase function of a lens whose focal length is determined by the relative orientation of the plates; hence, we could obtain a large aperture focus tunable lens by rotating one of the phase plates.
Other promising solutions to enable large-aperture tunable lensing include the Fresnel and  Pancharatnam-Berry liquid crystal lenses~\cite{jamali2018design,Jamali18} and tunable metasurface doublets~\cite{arbabi2018mems}.  
In all of these cases, our tracking method could be used to provide precise estimates of the focal length.

\subsubsection{Eye Box}
The eye box of multifocal displays are often small, and the proposed display is no exception. 
Due to the depth difference of focal planes, as the eye shifts, contents on each focal plane shift by different amounts, with the closer ones traverse more than the farther ones.  
This will leave uncovered as well as overlapping regions at depth discontinuities.  
	Further, the severity of the artifacts depends largely on the specific content being displayed. 
In practice, we observe that these artifacts are not distracting for small eye movements in the order of few millimeters. 
This problem can be solved by incorporating an eye tracker, as in~\citeN{mercier2017multifocal}.

\subsection{Reduced Maximum Brightness and Energy Efficiency}
Key limitations of our proposed design are the reduction in maximum brightness and, depending on the implementation, the energy efficiency of the device.
Suppose we are displaying $n$ focal planes per frame and $T$ frames per second.   
Each focal plane is displayed for $\frac{T}{n}$ second, which is $n$-times smaller compared to typical VR displays with one focal plane.
For our prototype, we use a high power LED to compensate for the reduction in brightness.
Further, brightness of the display is not a primary concern since there are no competing ambient lights sources for VR displays.

Energy efficiency of the proposed method  also depends on the type of display used.
For our prototype, since we use a DMD to spatially modulate the intensity at each pixel, we waste $\frac{n-1}{n}$ of the energy.
This can be completely avoided by adopted by using  OLED displays, where a pixel can be completely turned off.
An alternate solution is to use a phase spatial light modulator (SLM)~\cite{damberg2016high} to spatially redistribute a light source so that each focal plane only gets illuminated at pixels that need to be displayed; a challenge here is the slow refresh rate of the current crop of phase SLMs.
Another option is to use a laser along with a 2D galvo to selectively illuminate the content at each depth plane; however, 2D galvos are often slow when operated in non-resonant modes.

\begin{figure*}[th]
	\centering
	\begin{subfigure}[t]{0.65\linewidth}
		\centering
		\includegraphics[width=\linewidth]{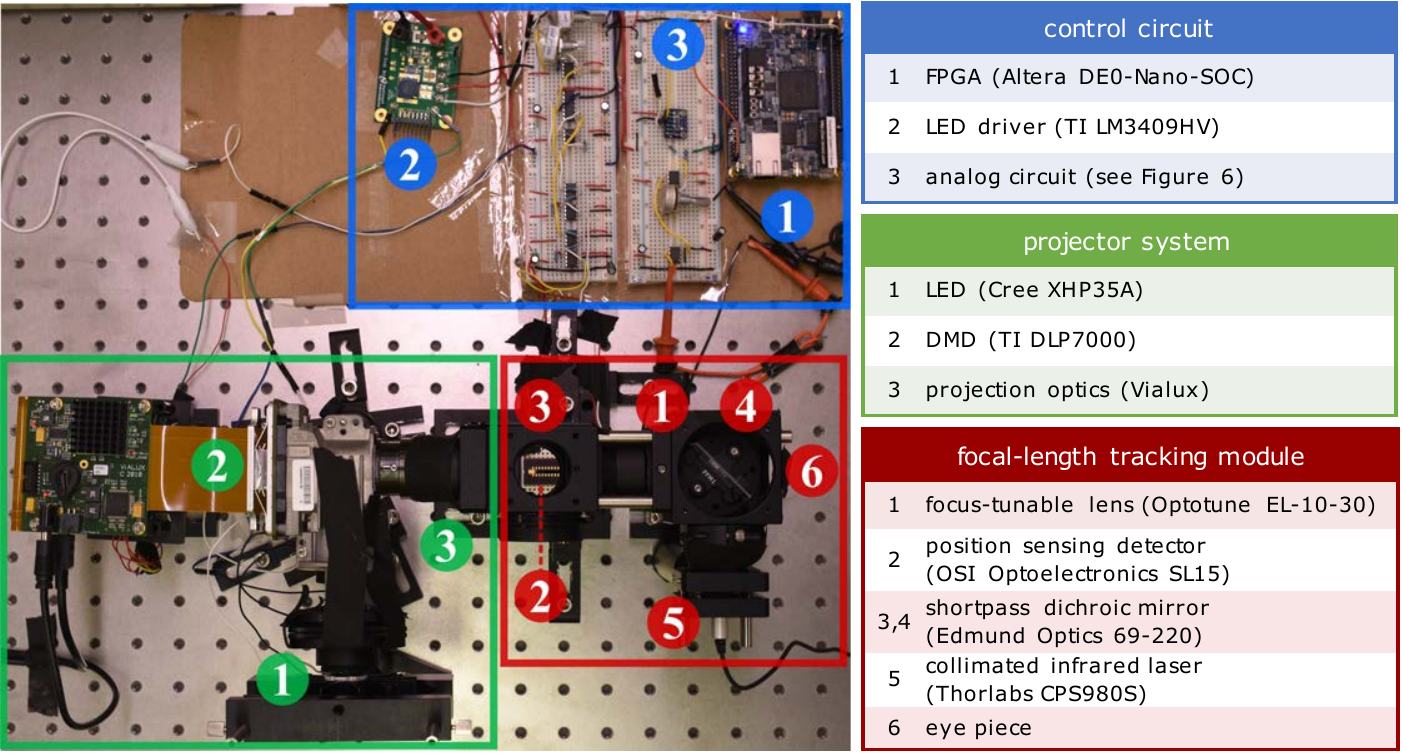}
		\vspace{-5mm}
		\caption{\footnotesize Photograph of the prototype and its component list}
	\end{subfigure}	
	~
	\begin{subfigure}[t]{0.34\linewidth}
		\centering
		\includegraphics[width=0.9\linewidth]{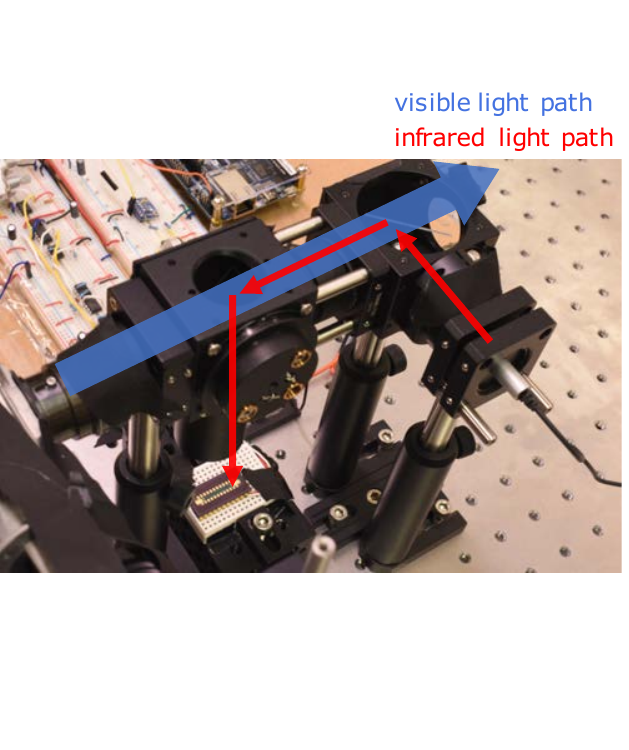}
		\vspace{-5mm}
		\caption{\footnotesize Light path for infrared and visible light}
	\end{subfigure}	
	\vspace{-3mm}
	\caption{The prototype is composed of a projector, the proposed focal-length tracking module, and the control circuits. (b) The two shortpass dichroic mirrors allow visible light to pass through and reflect infrared.  The enables us to create individual light path for each of them.}
	\label{figure: prototype}
\end{figure*}

\section{Proof-of-Concept Prototype}
\label{sec: prototype}

In this section, we present a lab prototype that  generates a dense focal stack using high-speed tracking of the focal length of a tunable lens and a high-speed display.

\subsection{Implementation Details}

The prototype is composed of three functional blocks:  the focus-tunable lens, the focal-length tracking device, and a DMD-based projector.
All the three components are controlled by an FPGA (Altera DE0-nano-SOC).
The FPGA drives the tunable lens with a digital-to-analog converter (DAC), following Algorithm~\ref{alg: control}.
Simultaneously, the FPGA reads the focal-length tracking output with an analog-to-digital converter (ADC) and uses the value to trigger the projector to display the next focal plane.
Every time a focal plane has been displayed, the projector is immediately turned off to avoid blur caused by the continuously changing focal-length configurations.
A photo of the prototype is shown in Figure~\ref{figure: prototype}.   
In the following, we will introduce each component in detail.

\subsubsection{Calibration.} 

In order to display focal planes at correct depths, we need to know the corresponding PSD tracking outputs.
From equations~\eqref{eq: 1_fx} and~\eqref{eq: Dx}, we have
\begin{equation}
\frac{1}{v(t)} = \frac{1}{d_o} - \frac{1}{d_p} - \frac{\ell}{2 a d_p} r(t) = \alpha + \beta r(t).
\label{eq: calibration}
\end{equation}
Thereby, we can estimate the current depth $v(t)$ if we know $\alpha$ and $\beta$, which only requires two measurements to estimate.
With a camera focused at $v_a = 25$ \cm and $v_b = \infty$, we get the two corresponding ADC readings $r_a$ and $r_b$.  
The two points can be accurately measured, since the depth-of-field of the camera at $25$ \cm is very small, and infinity can be approximated as long as the image is far away.
Since \eqref{eq: calibration} has an affine relationship, we only need to divide $[r_a, r_b]$ evenly into the desired number of focal planes.

\subsubsection{Control Algorithm.} 

The FPGA follows Algorithm~\ref{alg: control} to coordinate the tunable lens and the projector. 
On a high level, we drive the tunable lens with a triangular wave by continuously increasing/decreasing the DAC levels. 
We simultaneously detect the PSD's DAC reading $r$ to trigger the projection of focal planes.
When the last/first focal plane is displayed, we switch the direction of the waveform.
Note that while Algorithm~\ref{alg: control} is written in serial form, every module in the FPGA runs in parallel.

\begin{algorithm}[h]
	\KwData{$n$ target PSD triggers $r_1, \dots, r_n$}
	\KwIn{PSD ADC reading $r$}
	\KwOut{Tunable-lens DAC level $L$, projector display control signal}
	Initialize $L = 0$, $\Delta L = 1$, $ i = 1$\\
	\Repeat{manual stop}{
		
		$ L \gets L + \Delta L$  
		
		\If{$| r - r_i | \le \Delta r$}{
			Display focal plane $i$ and turn it off when finished.\\
			$i \gets i + \Delta L$\\
			\uIf{$\Delta L == 1$ and $i > n$}{
				Change triangle direction to down: 	$\Delta L \gets -1$, $i \gets n$
			}
			\uElseIf{$\Delta L == -1$ and $i < 1$}{
				Change triangle direction to up: \hspace{2.7mm}	$\Delta L \gets +1$, $i \gets 1$
			}
		}
	}
	\caption{Tunable-lens and focal-plane control}
	\label{alg: control}
\end{algorithm}

The control algorithm is simple yet robust.
It is known that the  transfer function of the tunable lens is sensitive to many factors, including device temperature and unexpected motor delay and errors~\cite{optotune}.  
In our experience, even with the same input waveform, we observe different offsets, peak-to-peak values on the PSD output waveform for each period.
Since the algorithm does not drive the tunable lens with fixed DAC values and instead directly detect the PSD output (\ie, the focal length of the tunable lens), it is robust to these unexpected factors.
However, the robustness comes with a price.  
Due to the motor delay, the peak-to-peak value $r_{\mbox{max}} - r_{\mbox{min}}$ is often a lot larger than $r_n - r_1$.
This causes the frame rate of the prototype ($1600$ focal planes per second, or $40$ focal planes per frame at $40$ fps) to be lower than the highest display frame rate ($2500$ focal planes per second).

Note that since 40 fps is close to the persistence of vision, our prototype sometimes leads to flickering. 
However, the capability of the proposed device is to increase the number of focal planes per second and as such we can get higher frame rate by trading off the focal planes per frame. 
For example, we can achieve 60 fps by operating at 26 focal planes per frame.

\begin{figure}[t]
	\centering
	\begin{subfigure}[t]{\linewidth}
		\centering
		\includegraphics[width=\linewidth]{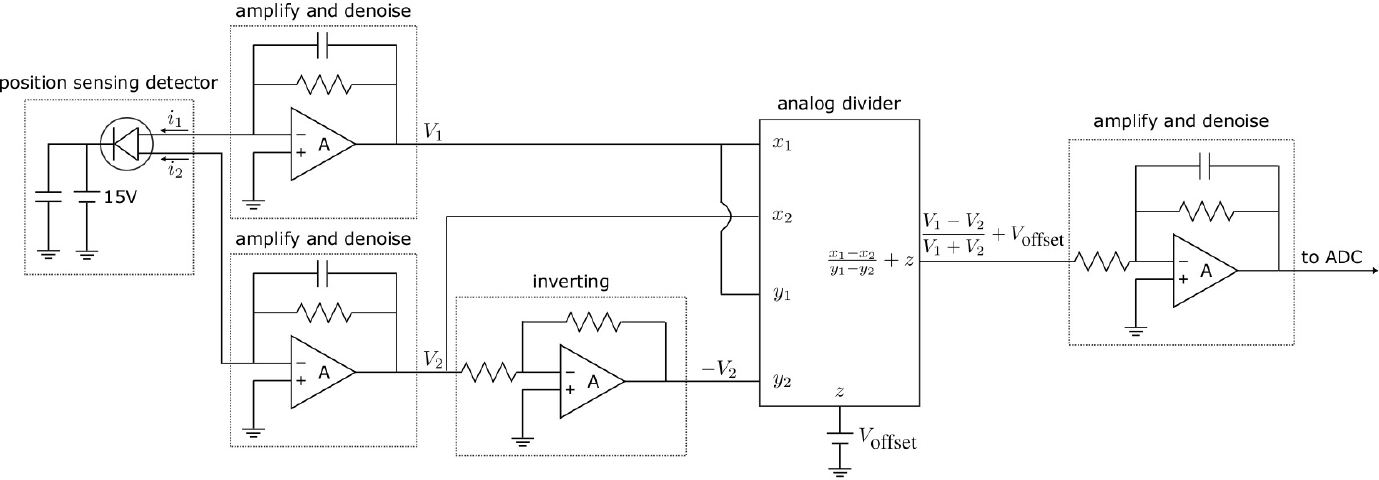}
		\vspace{-6mm}
		\caption{\footnotesize Analog circuit for processing focal-length tracking}
	\end{subfigure}
\\
	\vspace{2.5mm}
	\begin{subfigure}[t]{\linewidth}
		\centering
		\includegraphics[width=0.43\linewidth]{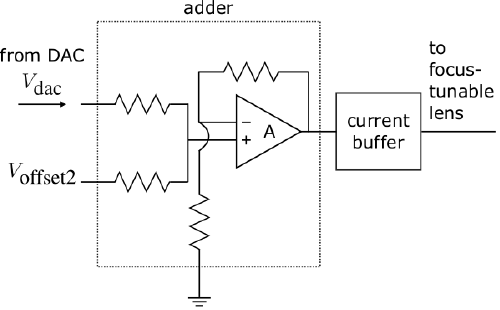}
		\vspace{-1.6mm}
		\caption{\footnotesize Analog circuit for driving focus-tunable lens}
	\end{subfigure}
	\vspace{-3mm}
	\caption{Analog circuits used in the prototype.  All the operational amplifiers are TI OPA-37, the analog divider is TI MPY634, and the current buffer is TI BUF634.  All denoising RC circuits have cutoff frequency at $47.7$ kHz.}
	\label{figure: circuit}
\end{figure}

\subsubsection{Focus-Tunable Lens and its Driver}
We use the focus-tunable lens EL-10-30 from Optotune~\cite{optotune}.  
The optical power of the lens ranges from approximately $8.3$ to $20$ diopters and is an affine function of the driving current input from $0$ to $300$~mA.
We use a 12-bit DAC (MCP4725) with a current buffer (BUF634) to drive the lens.  
The DAC provides $200$ thousand samples per second, and the current buffer has a bandwidth of $30$ MHz. 
This allows us to faithfully create a triangular input voltage up to several hundred Hertz.
The circuit is drawn in Figure~\ref{figure: circuit}b.

\subsubsection{Focal-Length Tracking and Processing}

The focal-length tracking device is composed of a one-dimensional PSD (SL15 from OSI Optoelectronics), two 800 \nm dichroic short-pass mirrors (Edmundoptics \#69-220), and a  980 \nm collimated infrared laser (Thorlabs CPS980S).
We drive the PSD with a reverse bias voltage of $15$~V.
This enables us to have $15$ \um precision on the PSD surface and rise time of $0.6$~\us. 
Across the designed accommodation range, the laser spot traverses within $7$ \mm on the PSD surface, which has a total length $15$ \mm.
This allows us to accurately differentiate up to $466$ focal-length configurations. 

The analog processing circuit has three stages --- amplifier, analog calculation, and an ADC, as shown in Figure~\ref{figure: circuit}a.
We use two operational amplifiers (TI OPA-37) to amplify the two output current of the PSD.
The gain-bandwidth of the amplifiers are $45$ MHz, which can fully support our desired operating speeds.
We also add a low-pass filter with a cut-off frequency of $47.7$ kHz at the amplifier, as a denoising filter.
The computation of $r(t)$ is conducted with two operational amplifiers (TI OPA-37) and an analog divider (TI MPY634).
We use a 12-bit ADC (LTC2308) with a  rate of $200$ thousand samples per second to port the analog voltage to the FPGA.  

Overall, the latency of the focal-length tracking circuit is $\sim\!20$ \us.
The bottleneck is the low-pass filter and the ADC; rest of the components have time responses in nanoseconds. 
Note that in $20$ \us the focal length of the tunable lens changes by $0.01$ diopters --- well below the detection capabilities of the eye~\cite{campbell1957depth}. 
Also, the stability of the acquired focal stack (which took a few hours to capture) indicates that the latency was either minimal or at least predictable and can be dealt with by calibration.

\subsubsection{DMD-based Projector}
The projector is composed of a DLP-7000 DMD from Texas Instruments, projection optics from Vialux, and a high-power LED XHP35A from Cree.  
We control the DMD with a development module Vialux V-7000. 
We update the configuration of micro-mirrors every $50$~\us.  
Following~\citeN{gray}, we use pulse-width modulation, performed through a LED driver (TI LM3409HV), to change the intensity of the LED concurrently with the update of micro-mirrors. 
This enables us to display at most $2500$ 8-bit images per second.
For simplicity, we preload each of the 40 focal planes onto the development module.   
Each focal stack requires $40 \times 8 = 320$ bitplanes, and thereby, we can store up to $136$ focal stacks on the module.  
The lack of video-streaming capability  needs further investigation to make it practical; it could potentially be resolved by using the customized display controller in~\citep{lincoln2016motion,lincoln2017} that is capable of displaying bitplanes with 80 \us latency.  
This would enable us to display $1562$ 8-bit focal planes per second.
We also note that whether we use depth filtering or not, the transmitted bitplanes are sparse since each pixel has content, at best, at a few depth planes. 
Thereby, we do not need to transmit the entire $320$ bitplanes.

Note that we divide the 8 bitplanes of each focal planes into two groups of 4 bitplanes, and we display the first group when the triangular waveform is increasing, and the other at the downward waveform.
From the results that will be presented in Section~\ref{sec: exp}, we can see that the images of the two groups align nicely.
This demonstrates the high accuracy of the focal-length tracking.

\begin{figure}[t]
	\centering
	\begin{subfigure}[t]{0.3\linewidth}
		\centering
		\includegraphics[width=\linewidth]{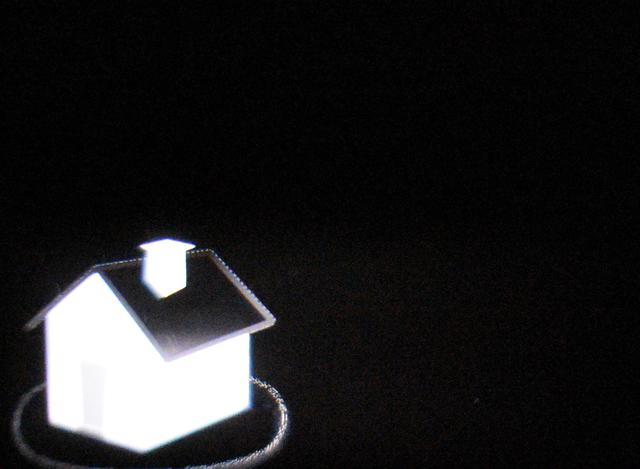}
	\end{subfigure}	
	~
	\begin{subfigure}[t]{0.3\linewidth}
		\centering
		\includegraphics[width=\linewidth]{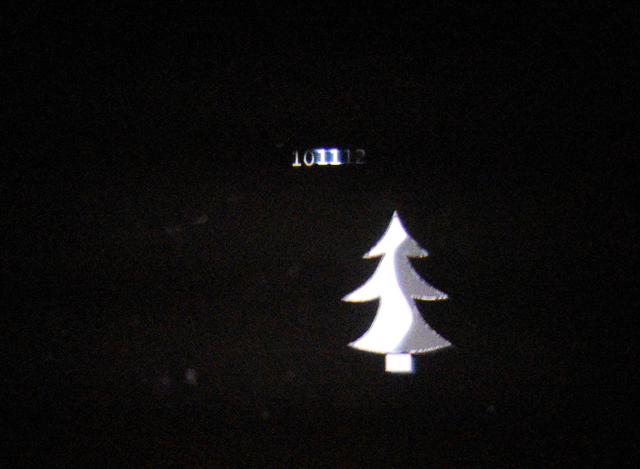}
	\end{subfigure}	
	~
	\begin{subfigure}[t]{0.3\linewidth}
		\centering
		\includegraphics[width=\linewidth]{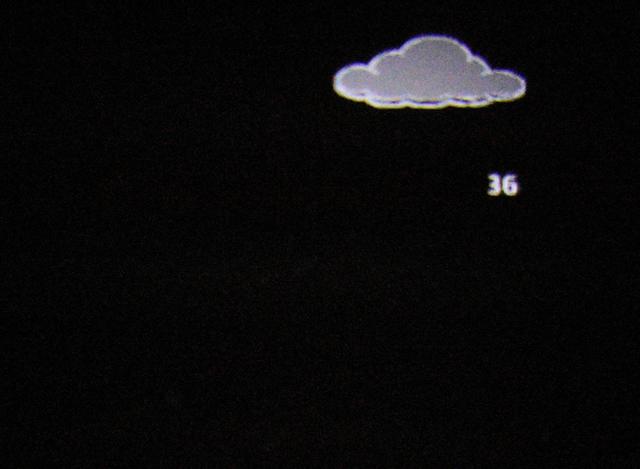}
	\end{subfigure}	
	\\
	\vspace{0.2mm}
		\begin{subfigure}[t]{0.3\linewidth}
		\centering
		\includegraphics[width=\linewidth]{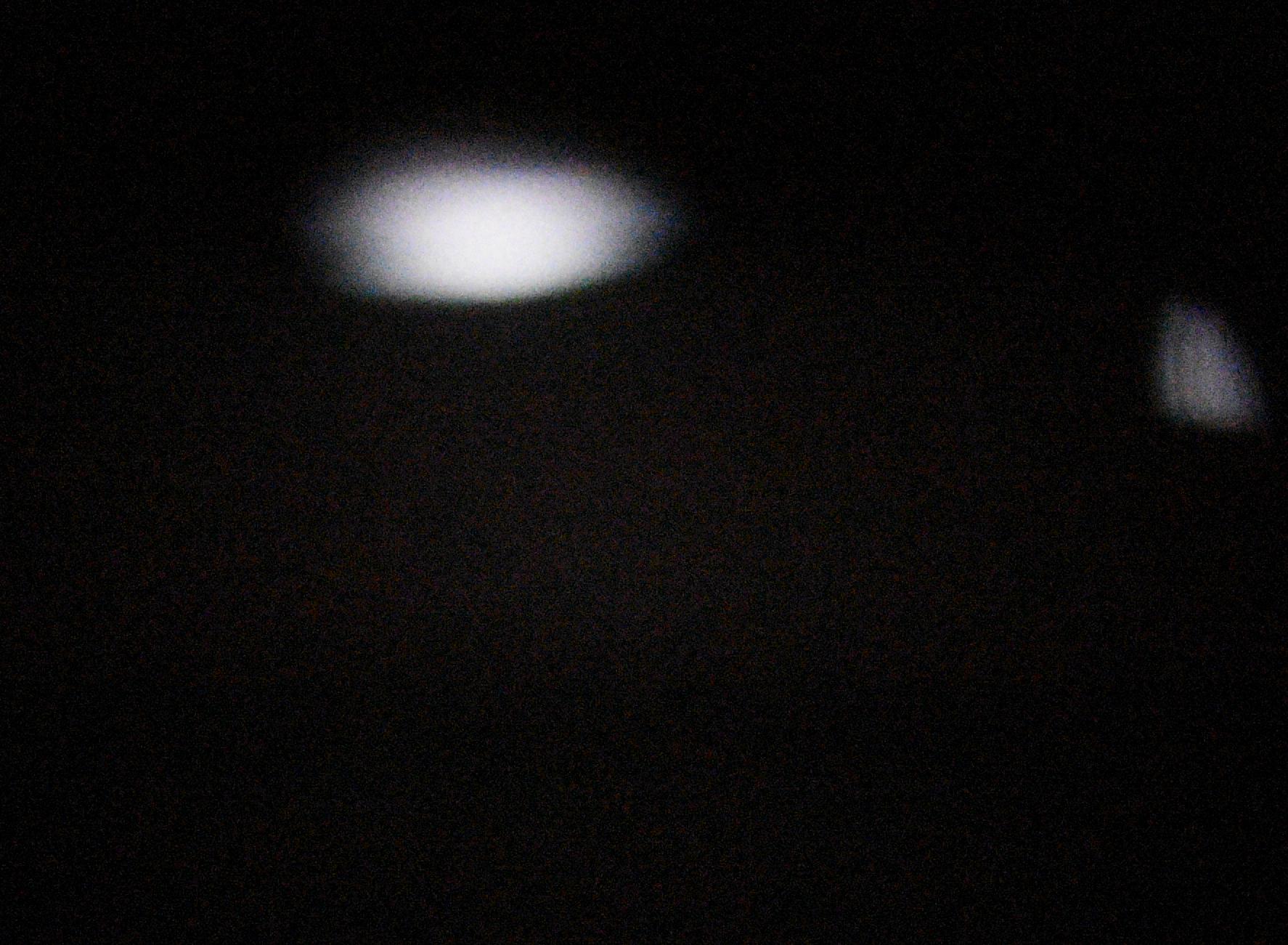}
	\end{subfigure}	
	~
	\begin{subfigure}[t]{0.3\linewidth}
		\centering
		\includegraphics[width=\linewidth]{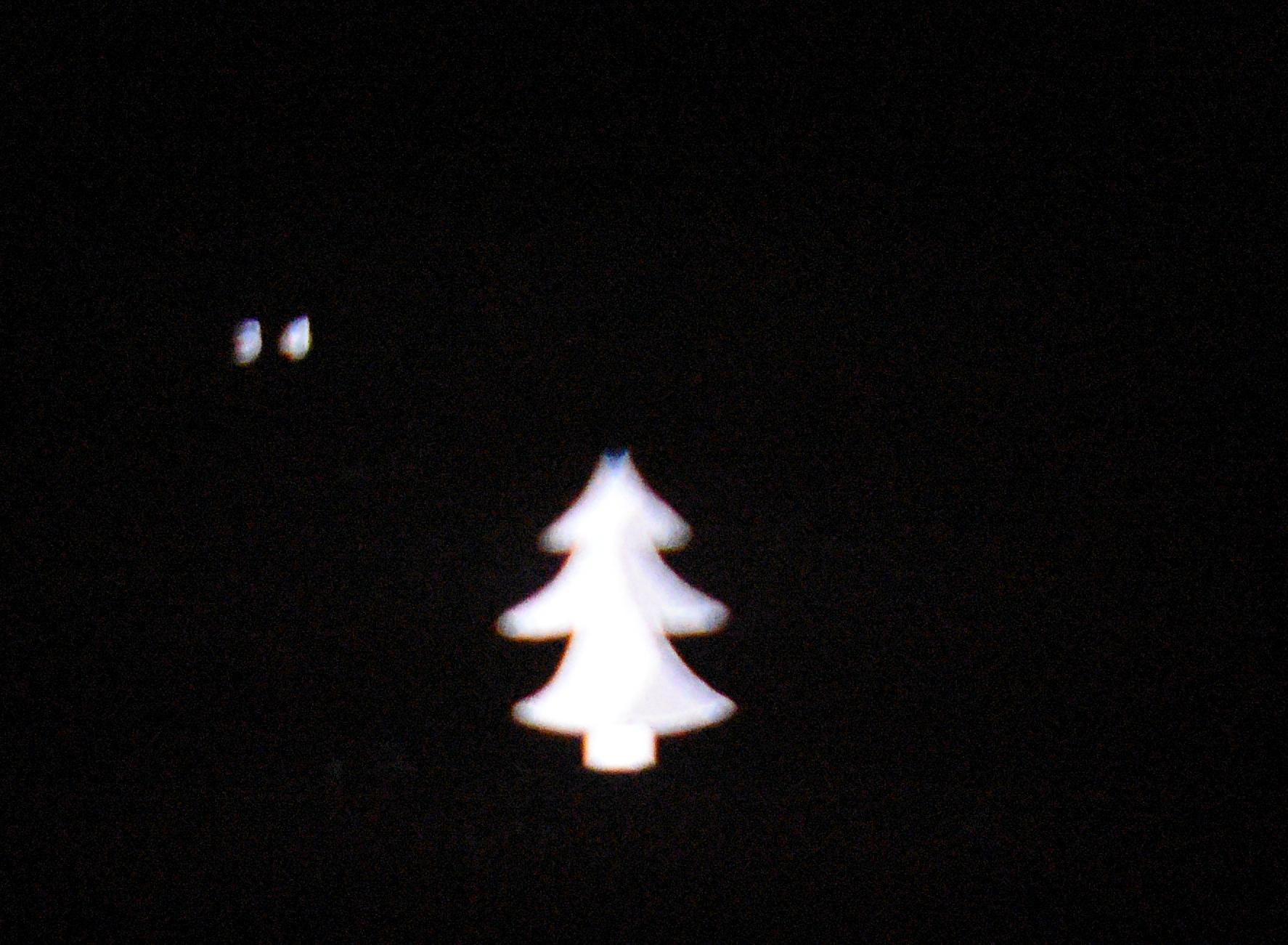}
	\end{subfigure}	
	~
	\begin{subfigure}[t]{0.3\linewidth}
		\centering
		\includegraphics[width=\linewidth]{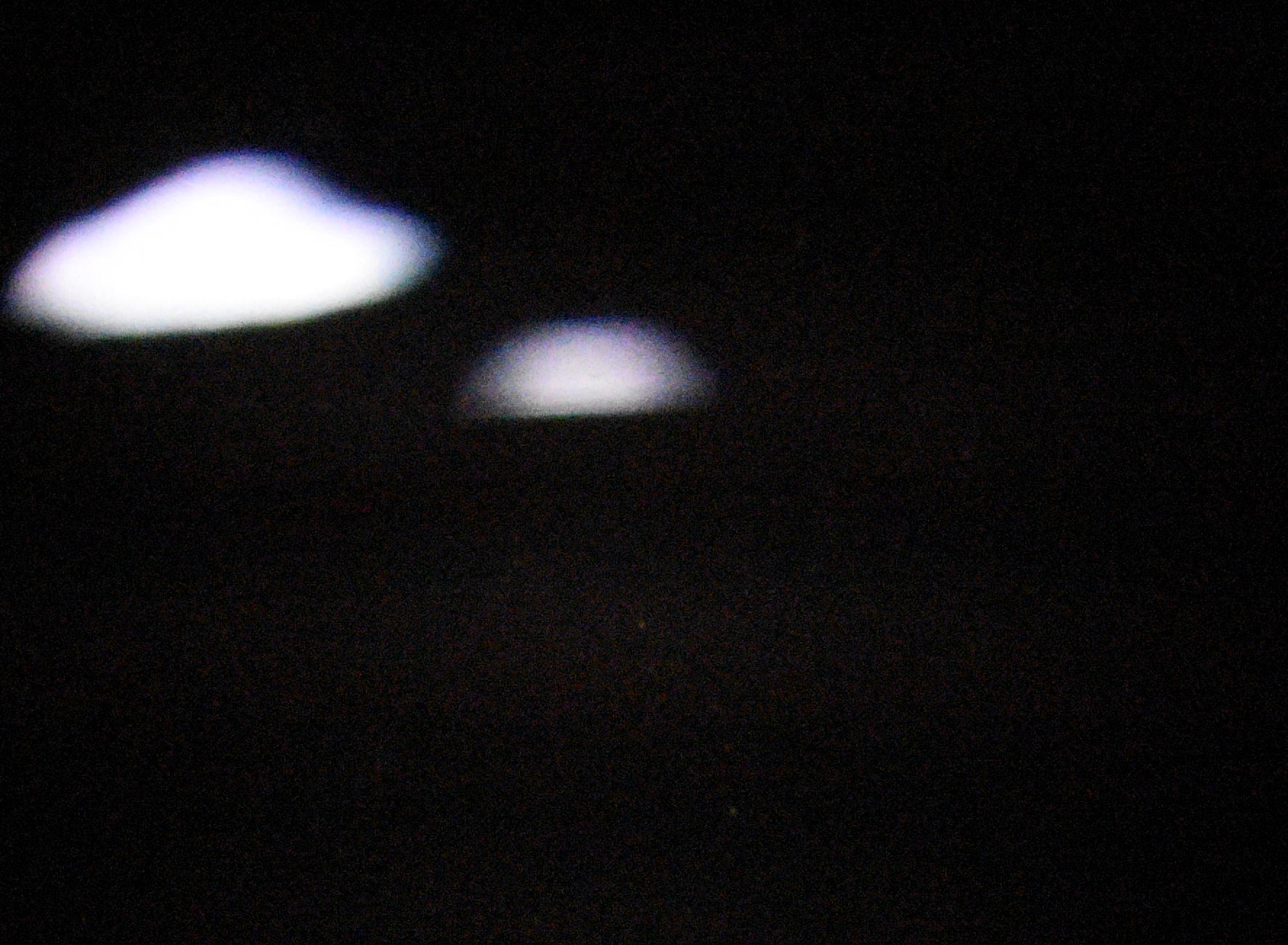}
	\end{subfigure}	
	\vspace{-3mm}
	\caption{Example images that are captured in burst shooting mode with a $f/4$ lens, exposure time equal to $0.5$ \ms, and ISO equal to $12,800$.   Note that in order to capture a single focal plane, we need exposure time of $0.2$ \ms.  Thereby, these images are composed of at most $3$ focal planes. }
	\label{figure: exp focal plane}
	\vspace{-3mm}
\end{figure}

As a quick verification of the prototype, we used the burst mode on the Nikon camera to capture multiple photographs at an aperture of $f/4$, ISO 12,800 and an exposure time of $0.5$ \ms.
Figure~\ref{figure: exp focal plane} shows six examples of displayed focal planes.  
Since a single focal plane requires an exposure time of $50 {\times} 4 =  0.2$ \ms, the captured images are  composed of at most $3$ focal planes.

\section{Experimental Evaluations}
\label{sec: exp}
We showcase the performance of our prototype on a range of scenes designed carefully to highlight the important features of our system.
The supplemental material has video illustrations that contain full camera focus stacks of all results in this section.

\subsection{Focal-Length Tracking}
\begin{figure}[t]
	\centering
	\includegraphics[width=\linewidth]{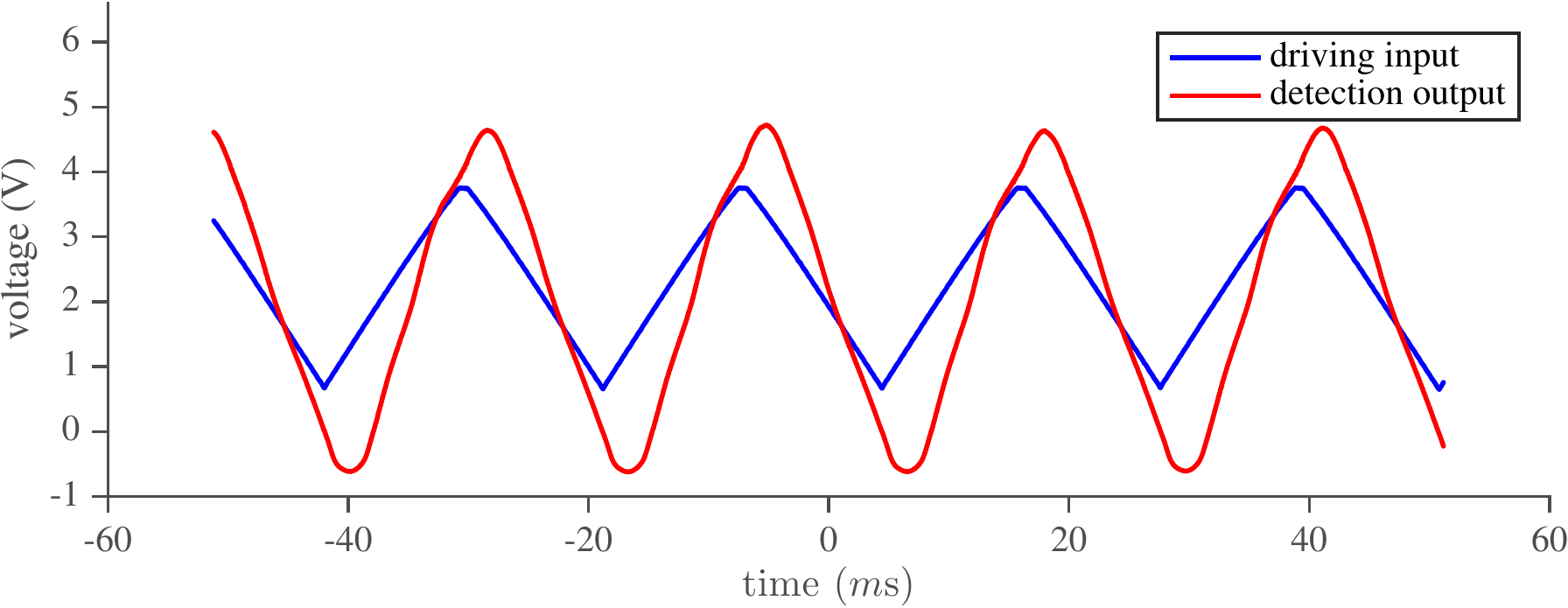}
	\vspace{-7mm}
	\caption{Measurements of the input signal to the tunable lens and the output of the PSD after analog processing.  The output waveform matches that of the input.  This shows that the proposed focal-length tracking is viable.}
	\label{figure: measurements}
\end{figure}

To evaluate the focal-length tracking module, we measure the input signal to the focus-tunable lens and the PSD output $r$ from an Analog Discovery oscilloscope.
The measurements are shown in Figure~\ref{figure: measurements}.
As can be seen, the output waveform matches that of the input. 
The high bandwidth of the PSD and the analog circuit enables us to track the focal length robustly in real-time.   
From the figure, we can also observe the delay of the focus-tunable lens ($\sim3$ \ms).

\subsection{ Depths of Focal Planes}
As stated previously, measuring the depth of the displayed focal planes is very difficult.  
Thereby, we use a method similar to depth-from-defocus to measure their depths.  
When a camera is focusing at infinity, the defocus blur kernel size will be linearly dependent on the depth of the (virtual) object in diopter.
This provides a method to measure the depths of the focal planes.

\begin{figure}[t]
	\centering
	\includegraphics[width=\linewidth]{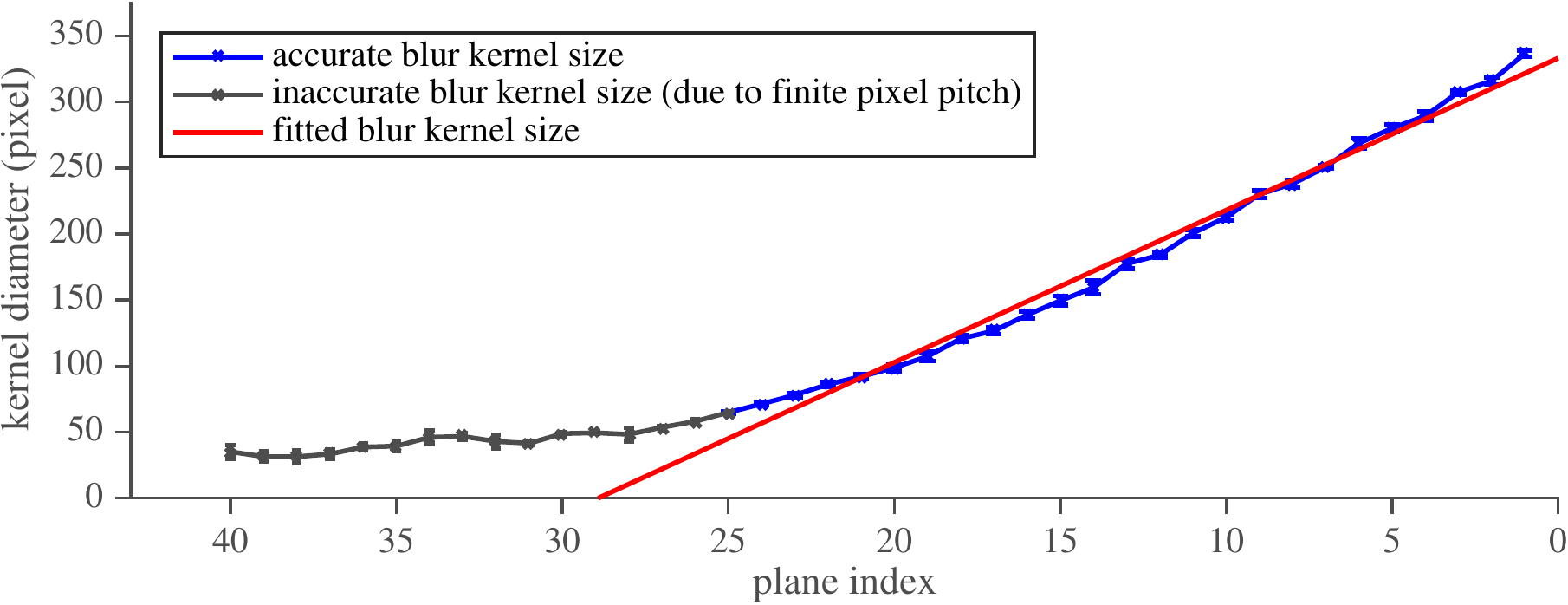}
	\vspace{-7mm}
	\caption{Measured blur kernel diameter by a camera focusing at infinity (plane $40$).  Due to the finite pixel pitch, the estimation becomes inaccurate when the spot size is too small (when the spots are displayed on focal planes close to infinity).  When the blur kernel size can be accurately estimated, they fit nicely as a linear segment. This indicates the depth of the focal planes are distributed uniformly in diopter.}
	\label{figure: blur kernel}
\end{figure}

For each of the focal plane, we display a $3 \times 3$ pixels white spot at the center, capture multiple images of various exposure time, and average the images to reduce noise.
We label the diameter of the defocus blur kernels and show the results in Figure~\ref{figure: blur kernel}.
As can be seen, when the blur-kernel diameters can be accurately estimated, \ie, largely defocus spots on closer focal planes, the values fit nicely to a straight line, indicating the depths of focal planes are uniformly separated in diopter.
However, as the  displayed spot size as a spot come into focus, the estimation of blur kernel diameters becomes inaccurate since we cannot display an infinitesimal spot due to the finite pixel pitch of the display.
Since there were no special treatments to individual planes in terms of system design or algorithm, we expect these focal planes to be placed accurately as well.

\subsection{Characterizing the System Point-Spread Function}
\begin{figure}[t]
	\centering
	\begin{subfigure}[t]{0.49\linewidth}
		\centering
		\includegraphics[width=\linewidth]{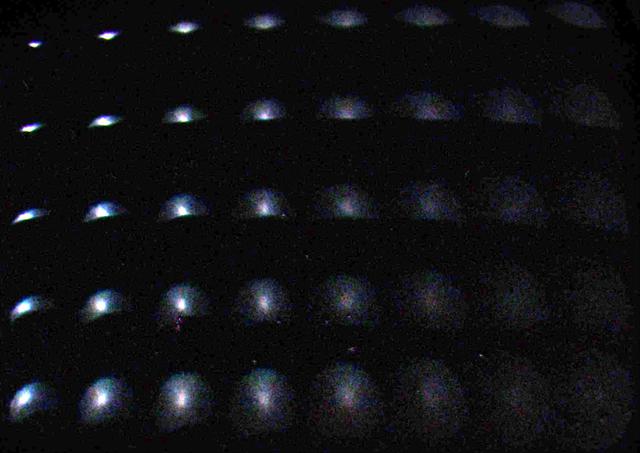}
		\vspace{-5mm}
		\caption{\footnotesize Camera focuses at  $25$ \cm}
	\end{subfigure}	
	\begin{subfigure}[t]{0.49\linewidth}
		\centering
		\includegraphics[width=\linewidth]{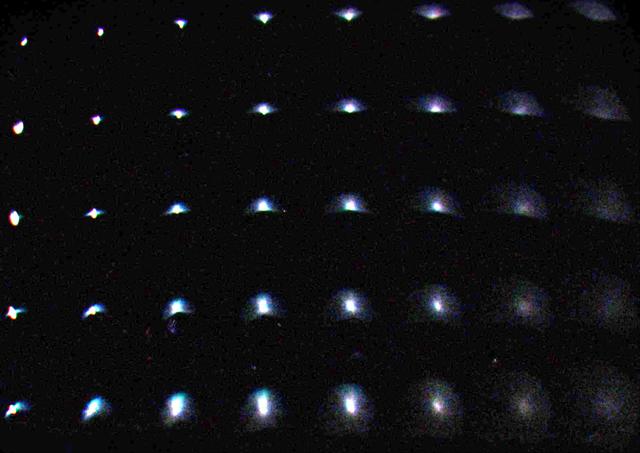}
		\vspace{-5mm}
		\caption{\footnotesize Camera focuses at  $30$ \cm}
	\end{subfigure}	
	\\
	\vspace{1.6mm}
	\begin{subfigure}[t]{0.49\linewidth}
		\centering
		\includegraphics[width=\linewidth]{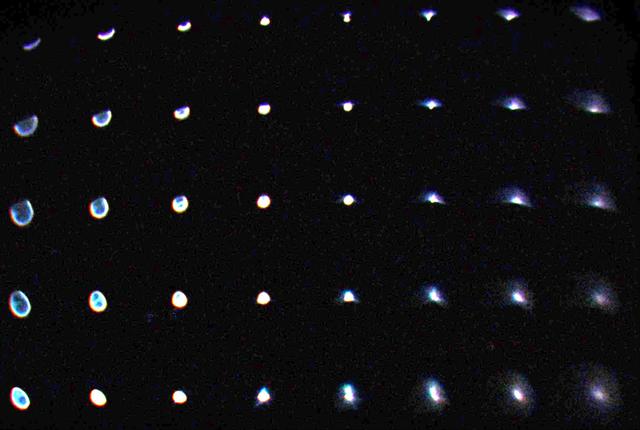}
		\vspace{-5mm}
		\caption{\footnotesize Camera focuses at  $\sim1.2$ m}
	\end{subfigure}	
	~
	\begin{subfigure}[t]{0.49\linewidth}
		\centering
		\includegraphics[width=\linewidth]{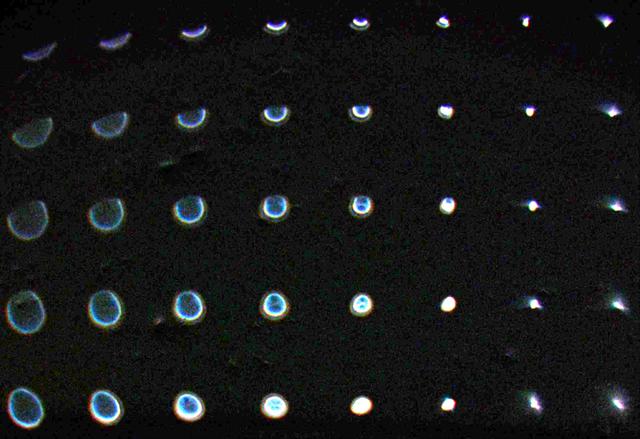}
		\vspace{-5mm}
		\caption{\footnotesize Camera focuses at infinity }
	\end{subfigure}	
	\vspace{-3mm}	
	\caption{Measured point spread function of the prototype.  Each of the $40$ points is on a different focal plane --- the top-left is closest to the camera and the bottom-right is farthest.  For better visualization, we multiply the image by $10$ and filter the image with a $4\times4$ median filter.  The results show that the prototype is able to produce $40$ distinct focal planes.}
	\label{fig:psf}
\end{figure}

To characterize our prototype, we measure its point spread function with a Nikon D3400 using a $50$ \mm $f/1.4$ prime lens.
We display a static scene that is composed of $40$ $3\times3$ spots with each spot at a different focal plane.
Using the camera, we capture a focal stack of $169$ images ranging from $0$ to $4$ diopters away from the focus-tunable lens.
For improved contrast, we remove the background and noise due to dust and scratches on the lens by capturing the same focal stack with no spot shown on the display.
Figure~\ref{fig:psf} shows the point spread function of the display at four different focus settings, and a video of this focal stack is attached in the supplemental material.
The result shows that the prototype is able to display the spots at $40$ depths concurrently within a frame, verifies the functionality of the proposed method.
The shape and the asymmetry of the blur kernels can be attributed to the spherical aberration of the focus-tunable lens as well as the throw of the projection lens on the DMD.

\def \one {gt-region1}
\def \onedes {ground truth}
\def \two {direct4-region1}
\def \twodes {4-plane, direct}
\def \three {linear4-region1}
\def \threedes {4-plane, linear filtering}
\def \four {opt4-region1}
\def \fourdes {4-plane, opt filtering}
\def \five {direct40-region1}
\def \fivedes {40-plane, direct}
\def \six {opt40-region1}
\def \sixdes {40-plane, opt filtering}

\begin{figure*}[t]
	\centering
	\def \folder {e_1}
	\begin{subfigure}[t]{0.36\linewidth}
		\centering
		\begin{subfigure}[t]{0.32\linewidth}
			\centering
			\includegraphics[width=\linewidth]{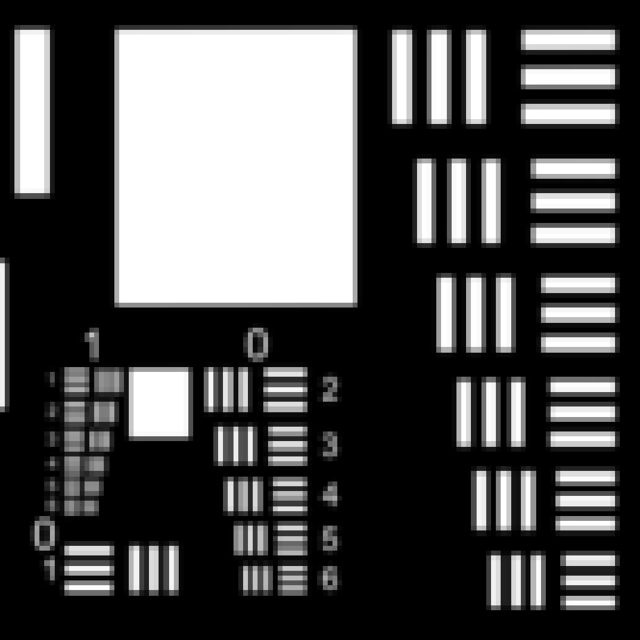}
			\vspace{-6mm}
			\caption*{\scriptsize \onedes}
		\end{subfigure}	
		\begin{subfigure}[t]{0.32\linewidth}
			\centering
			\includegraphics[width=\linewidth]{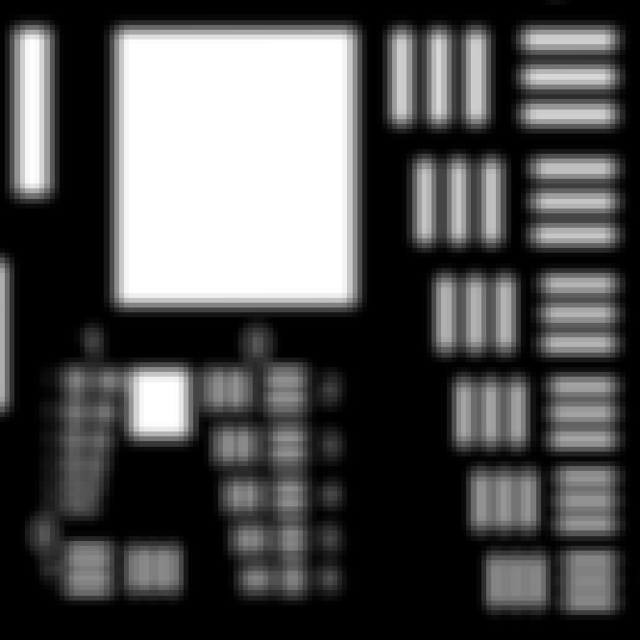}
			\vspace{-6mm}
			\caption*{\scriptsize \twodes}
		\end{subfigure}	
		\begin{subfigure}[t]{0.32\linewidth}
			\centering
			\includegraphics[width=\linewidth]{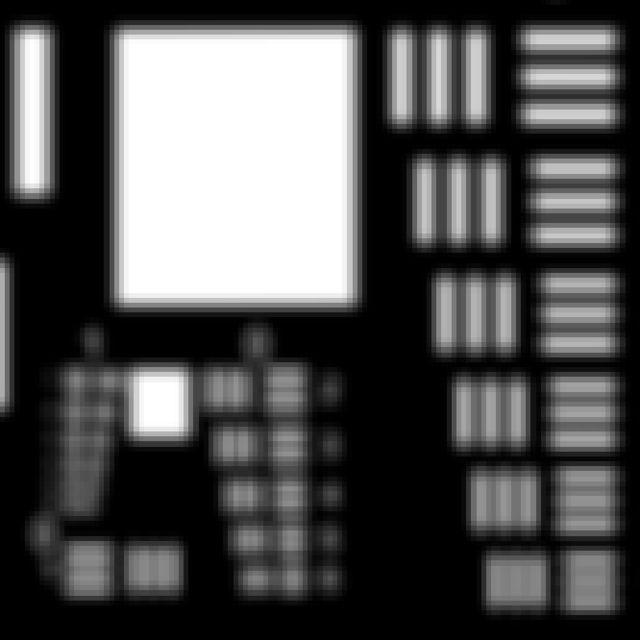}
			\vspace{-6mm}
			\caption*{\scriptsize \threedes}
		\end{subfigure}	
		\vspace{1mm}
		\\
		\begin{subfigure}[t]{0.32\linewidth}
			\centering
			\includegraphics[width=\linewidth]{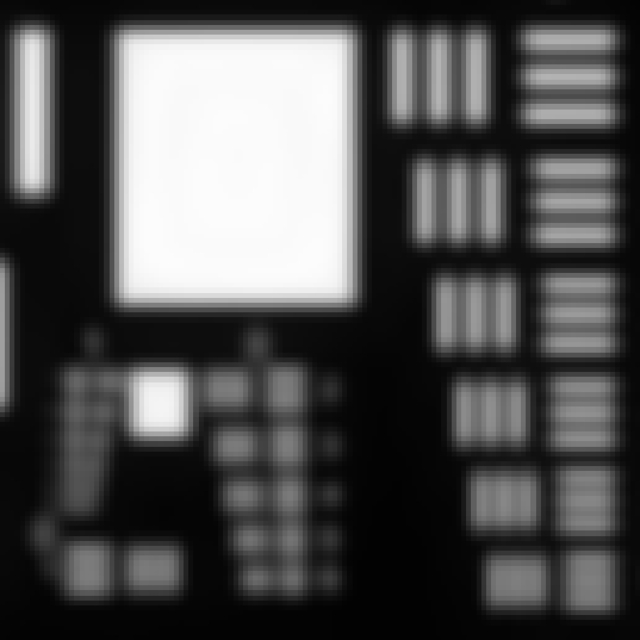}
			\vspace{-6mm}
			\caption*{\scriptsize \fourdes}
		\end{subfigure}	
		\begin{subfigure}[t]{0.32\linewidth}
			\centering
			\includegraphics[width=\linewidth]{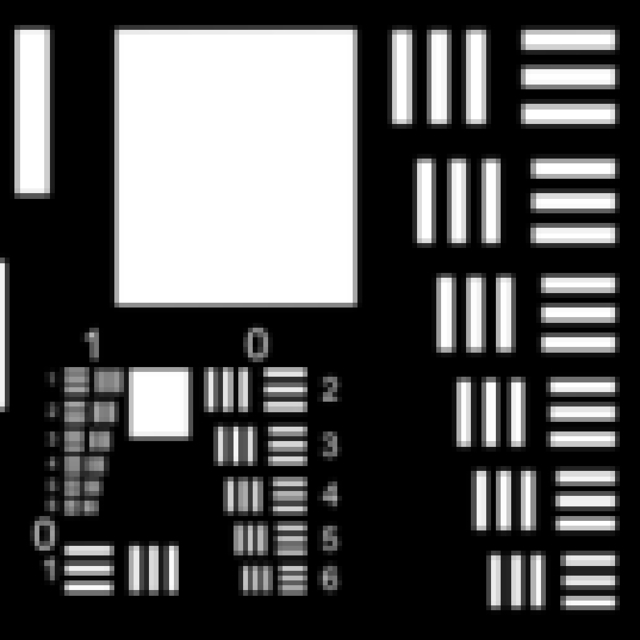}
			\vspace{-6mm}
			\caption*{\scriptsize \fivedes}
		\end{subfigure}	
		\begin{subfigure}[t]{0.32\linewidth}
			\centering
			\includegraphics[width=\linewidth]{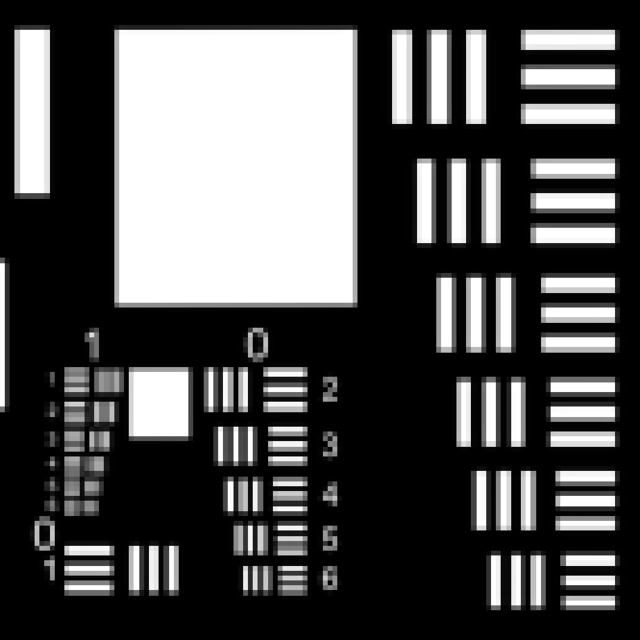}
			\vspace{-6mm}
			\caption*{\scriptsize \sixdes}
		\end{subfigure}	
		\vspace{-1mm}
		\caption{\footnotesize Camera focuses at $0.02$ diopters}
	\end{subfigure}	
	\hspace{1mm}
	\def \folder {e_19}
	\begin{subfigure}[t]{0.36\linewidth}
	\centering
		\begin{subfigure}[t]{0.32\linewidth}
			\centering
			\includegraphics[width=\linewidth]{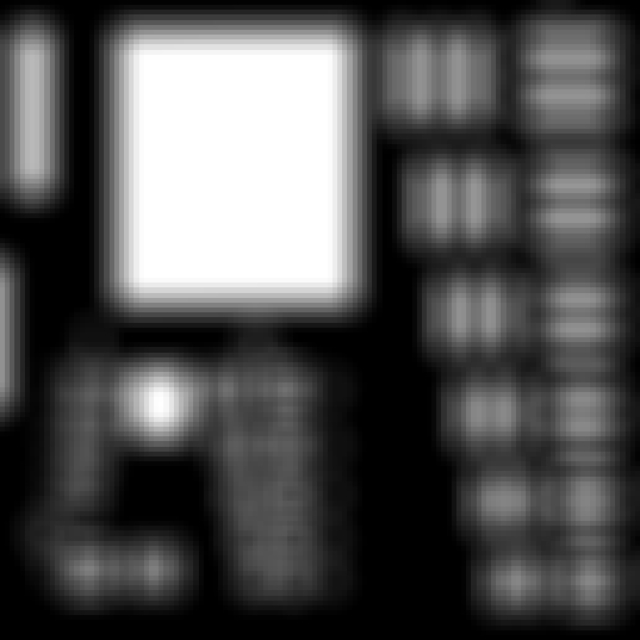}
			\vspace{-6mm}
			\caption*{\scriptsize \onedes}
		\end{subfigure}	
		\begin{subfigure}[t]{0.32\linewidth}
			\centering
			\includegraphics[width=\linewidth]{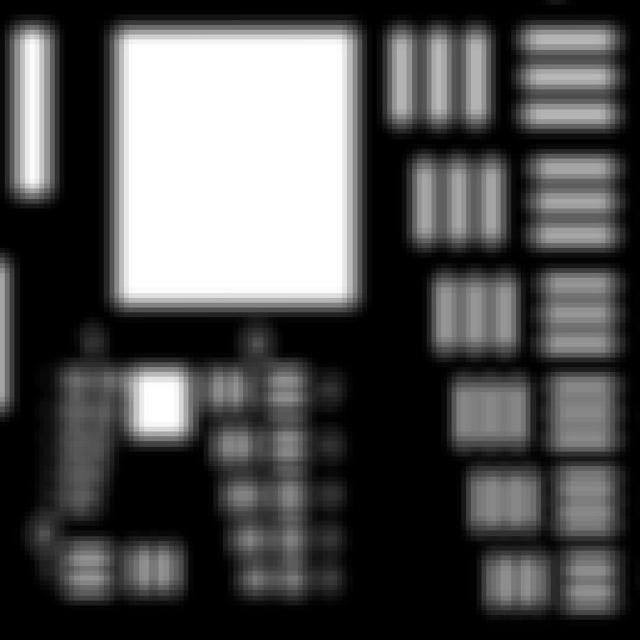}
			\vspace{-6mm}
			\caption*{\scriptsize \twodes}
		\end{subfigure}	
		\begin{subfigure}[t]{0.32\linewidth}
			\centering
			\includegraphics[width=\linewidth]{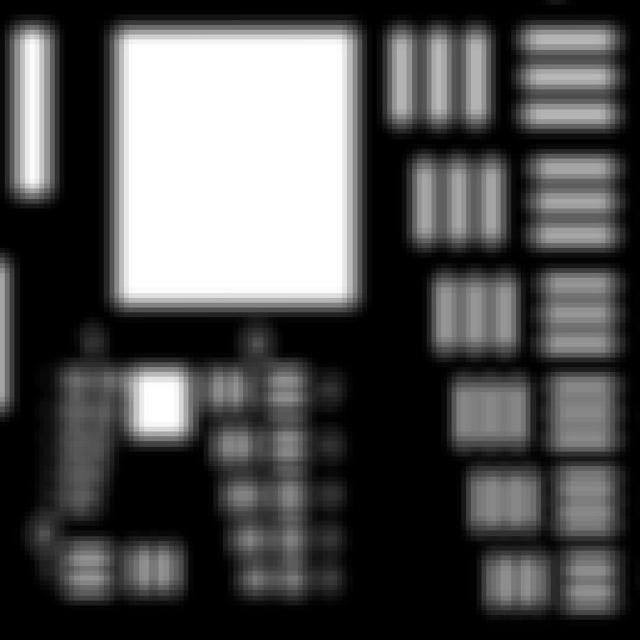}
			\vspace{-6mm}
			\caption*{\scriptsize \threedes}
		\end{subfigure}	
		\vspace{1mm}
		\\
		\begin{subfigure}[t]{0.32\linewidth}
			\centering
			\includegraphics[width=\linewidth]{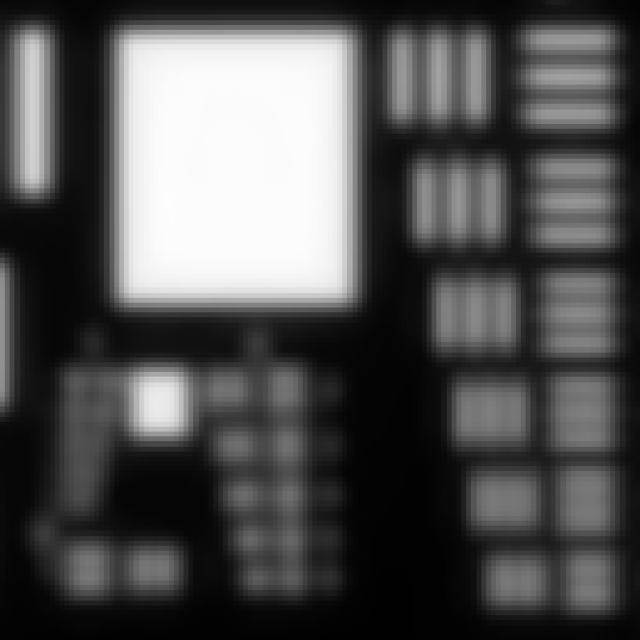}
			\vspace{-6mm}
			\caption*{\scriptsize \fourdes}
		\end{subfigure}	
		\begin{subfigure}[t]{0.32\linewidth}
			\centering
			\includegraphics[width=\linewidth]{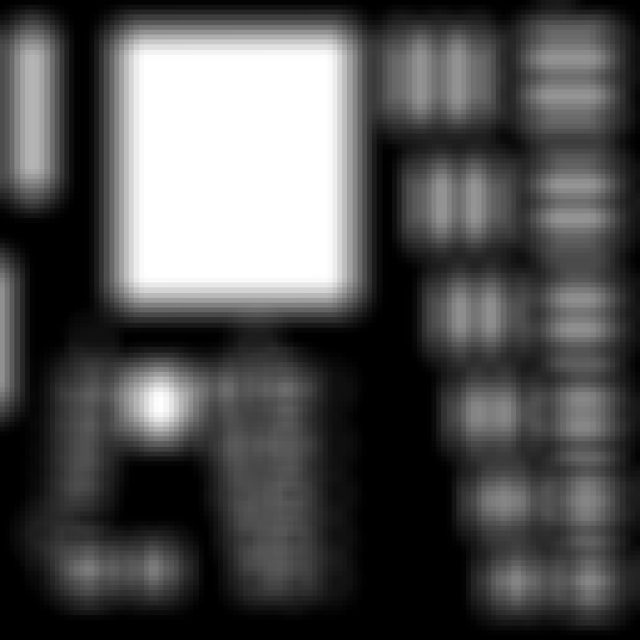}
			\vspace{-6mm}
			\caption*{\scriptsize \fivedes}
		\end{subfigure}	
		\begin{subfigure}[t]{0.32\linewidth}
			\centering
			\includegraphics[width=\linewidth]{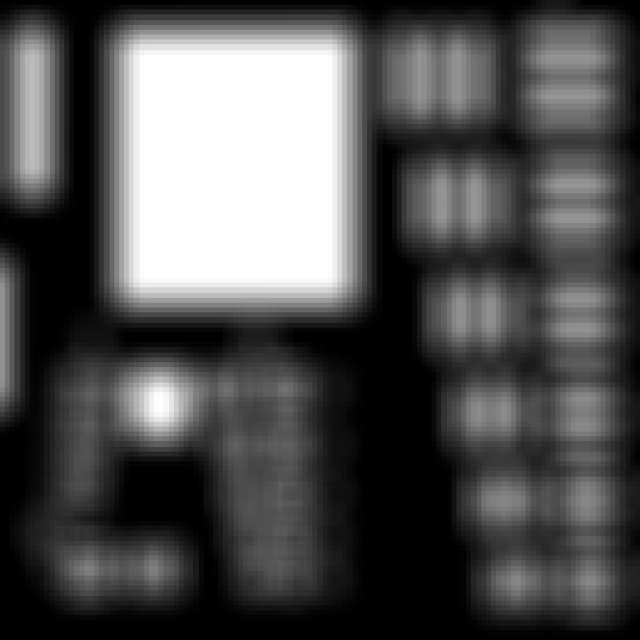}
			\vspace{-6mm}
			\caption*{\scriptsize \sixdes}
		\end{subfigure}	
		\vspace{-1mm}
		\caption{\footnotesize Camera focuses at $0.9$ diopters}
	\end{subfigure}
	\def \folder {e_19}
	\begin{subfigure}[c]{0.24\linewidth}
		\centering
		\vspace{16mm}
		\includegraphics[width=\linewidth]{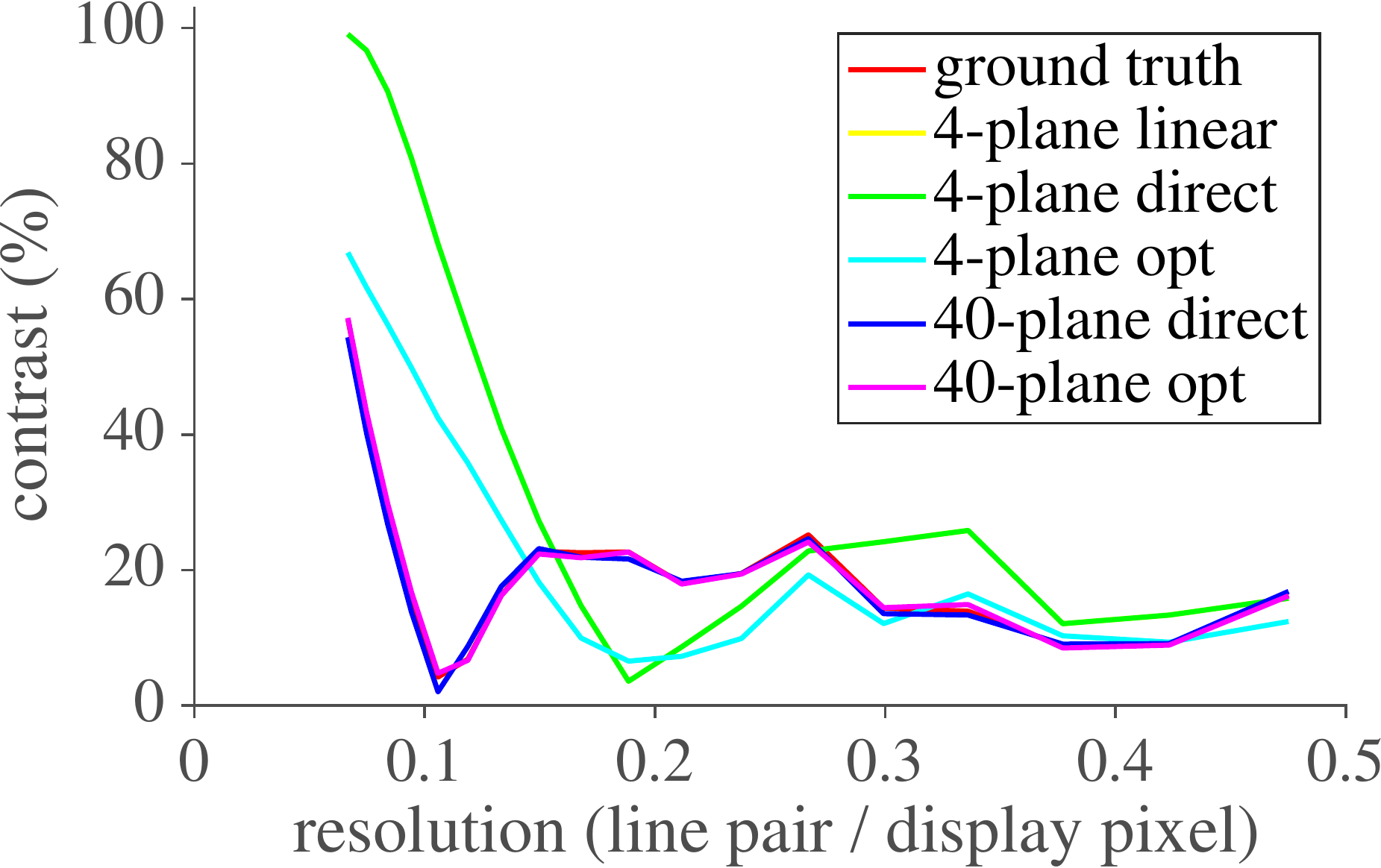}
		\caption{\footnotesize Modulation transfer functions of (b)}
	\end{subfigure}		
	\vspace{-2mm}
	\caption{Simulation results of 4-plane and 40-plane multifocal displays with direct quantization, linear depth filtering, and optimization-based filtering.  The scene is at $0.02$ diopters, which is an inter-plane location of the 4-plane display.  (a) When the camera focuses at $0.02$ diopters, the 40-plane display achieves higher spatial resolution than the 4-plane display, regardless of the depth filtering algorithm.  (b) When the camera focuses at $0.9$ diopters, the defocus blur on the 40-plane display closely follows that of the ground truth, whereas the 4-plane display fails to blur the low frequency contents.  This can also be seen from the modulation transfer function plotted in (c).  
	}
	\label{figure: res results}
\end{figure*}	 

\begin{figure*}[t]
	\centering
	\begin{subfigure}[t]{0.145\linewidth}
		\centering
		\includegraphics[width=\linewidth]{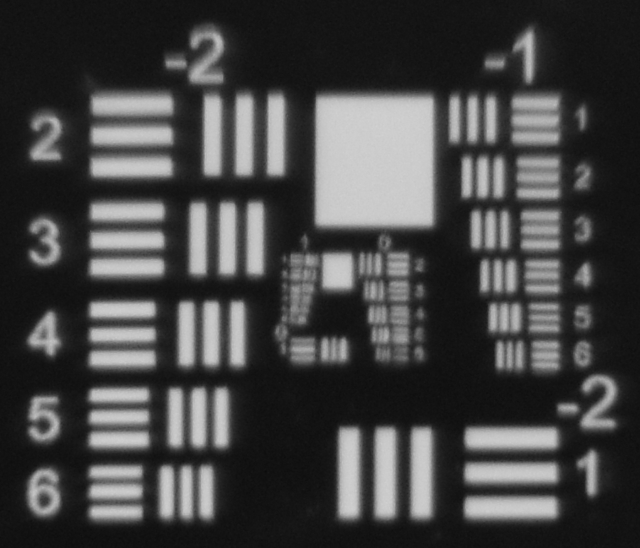}
		\vspace{-5mm}
		\caption{\footnotesize in-focus}
	\end{subfigure}	
	\begin{subfigure}[t]{0.145\linewidth}
		\centering
		\includegraphics[width=\linewidth]{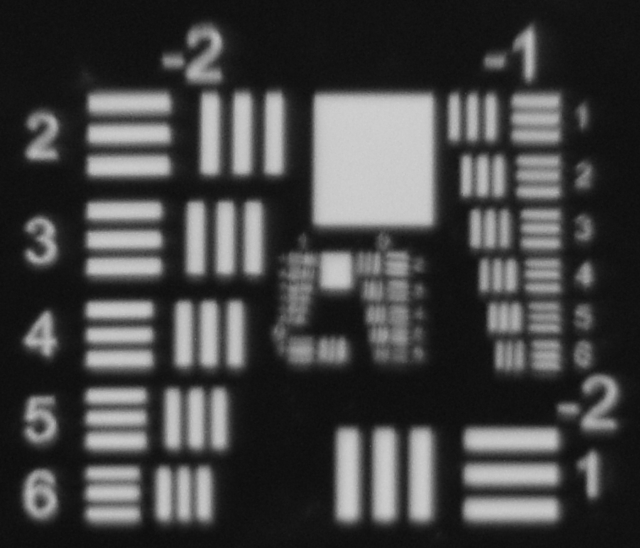}
		\vspace{-5mm}
		\caption{\footnotesize 40-plane}
	\end{subfigure}	
	\begin{subfigure}[t]{0.145\linewidth}
		\centering
		\includegraphics[width=\linewidth]{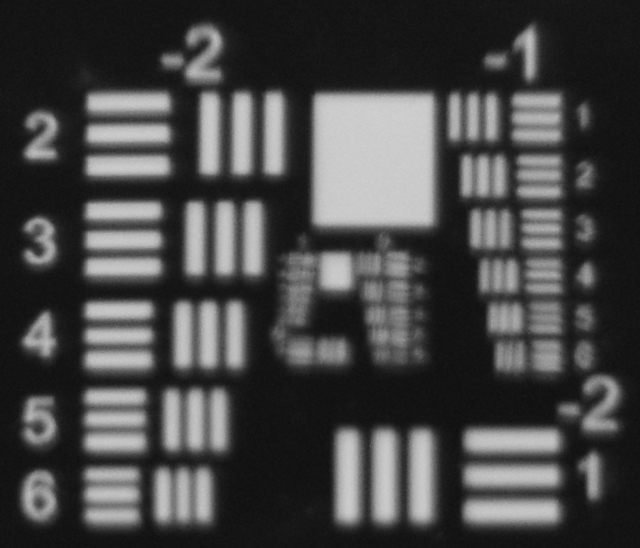}
		\vspace{-5mm}
		\caption{\footnotesize 30-plane}
	\end{subfigure}	
	\begin{subfigure}[t]{0.145\linewidth}
		\centering
		\includegraphics[width=\linewidth]{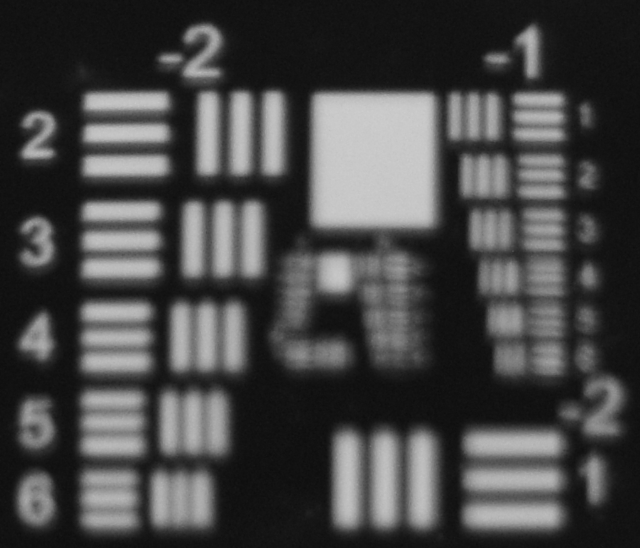}
		\vspace{-5mm}
		\caption{\footnotesize 20-plane}
	\end{subfigure}	
	\begin{subfigure}[t]{0.145\linewidth}
		\centering
		\includegraphics[width=\linewidth]{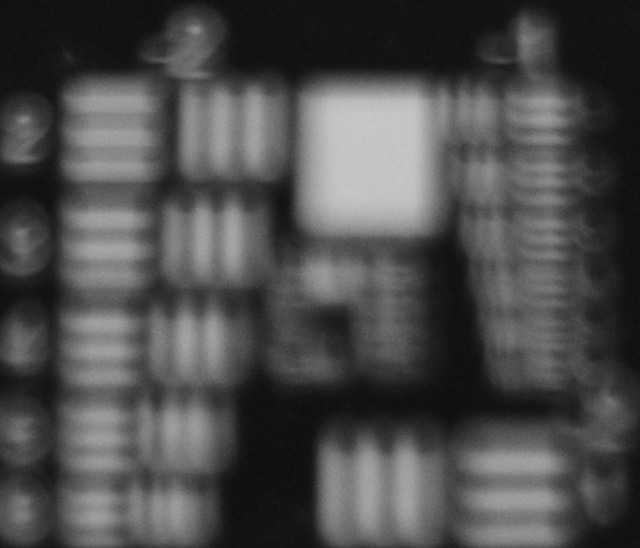}
		\vspace{-5mm}
		\caption{\footnotesize 4-plane}
	\end{subfigure}	
	\begin{subfigure}[t]{0.24\linewidth}
		\centering
		\includegraphics[width=\linewidth]{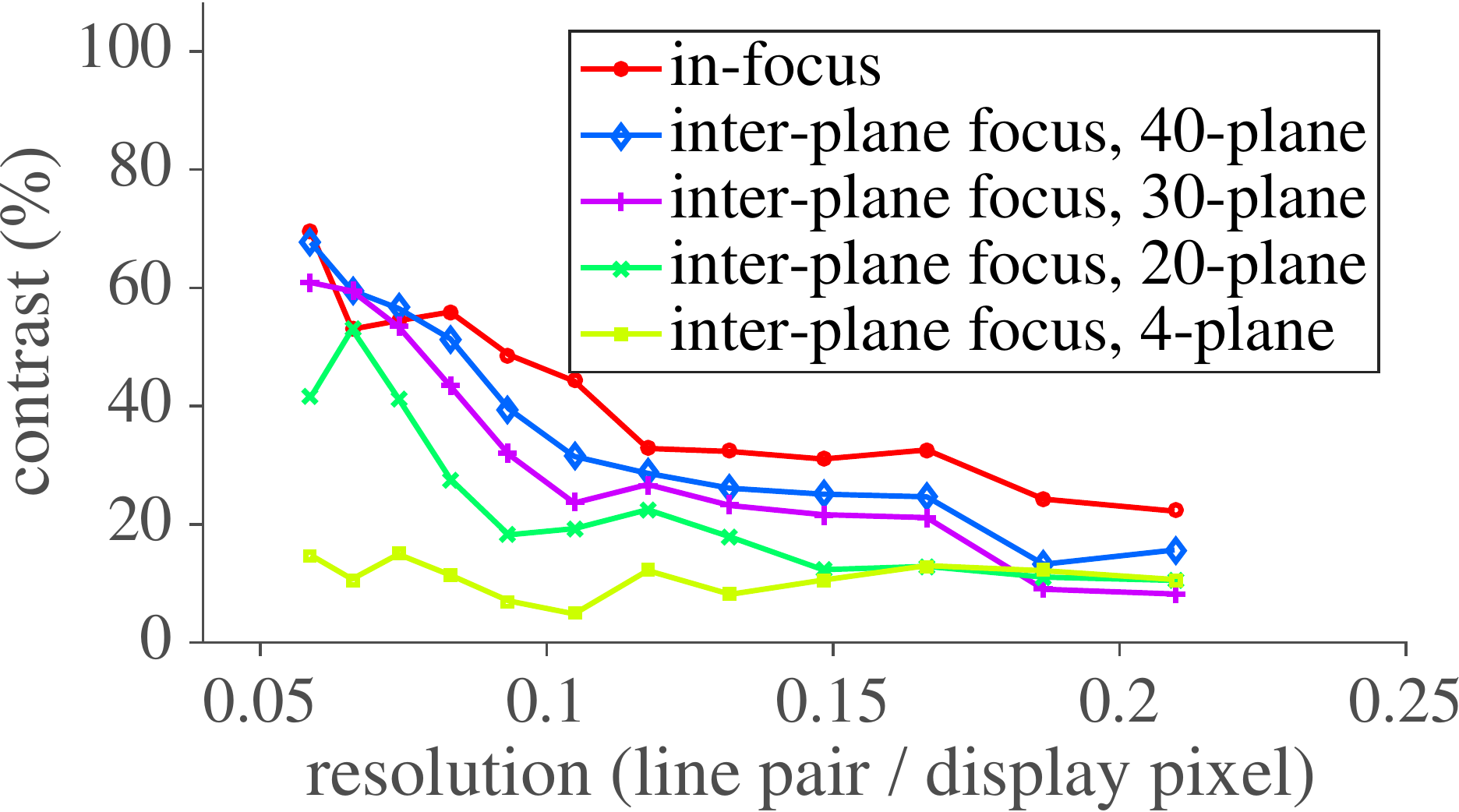}
		\vspace{-5mm}
		\caption{\footnotesize modulation transfer function}
	\end{subfigure}	
	\vspace{-2mm}
	\caption{Captured inter-plane focused images using a $50$ \mm $f/1.4$ lens.   The resolution chart locates on the 5th focal plane of the 40-plane display.  We emulate a 4-plane and a 20-plane display by putting their focal planes on the $5,15,25,35$th and on the odd focal planes of the 40-plane display, respectively.
		(a) Camera focuses at the 5th focal plane.   (b,c) Cameras focus at the estimated inter-plane locations of the 40-plane display and the 30-plane displays, respectively.  (d) Camera focuses at the 6th focal plane, an inter-plane location of a 20-plane display.  (e) Camera focuses at the 10th focal plane, an inter-plane location of a 4-plane display.   Their modulation transfer functions are plotted in (f).  
	}
	\label{figure: interplane}
\end{figure*}

\def\figurehieght{0.75\linewidth}
\begin{figure*}[t]
	\centering
	\begin{subfigure}[t]{\linewidth}
		\centering
		\begin{subfigure}[t]{0.24\linewidth}
			\centering
			\caption*{\footnotesize focused at the 1st focal plane ($0.25$ m)}
		\end{subfigure}	
		\hspace{0.5mm}
		\begin{subfigure}[t]{0.24\linewidth}
			\centering
			\caption*{\footnotesize focused at the 10th focal plane ($0.32$ m)}
		\end{subfigure}	
		\hspace{0.3mm}
		\begin{subfigure}[t]{0.24\linewidth}
			\centering
			\caption*{\footnotesize focused at the 20th focal plane ($0.49$ m)}
		\end{subfigure}	
		\begin{subfigure}[t]{0.24\linewidth}
			\centering
			\caption*{\footnotesize focused at the 40th focal plane at $\infty$ }
		\end{subfigure}	
	\end{subfigure}
	\\
	\vspace{0.5mm}
	\def\dataset{4d}
	\begin{subfigure}[t]{\linewidth}
		\centering
		\begin{subfigure}[t]{0.24\linewidth}
			\centering
			\includegraphics[height=\figurehieght]{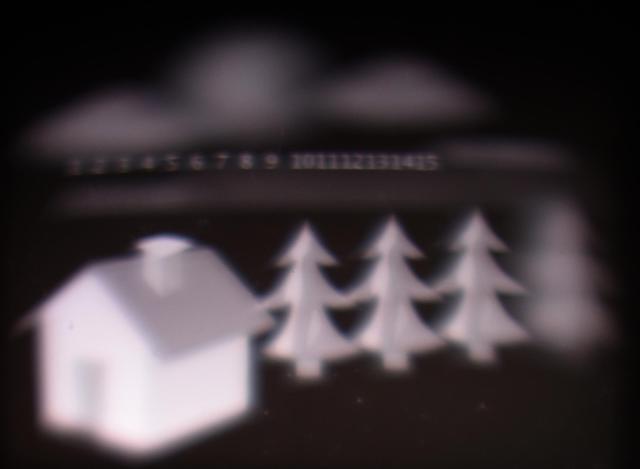}
		\end{subfigure}	
		\hspace{0.5mm}
		\begin{subfigure}[t]{0.24\linewidth}
			\centering
			\includegraphics[height=\figurehieght]{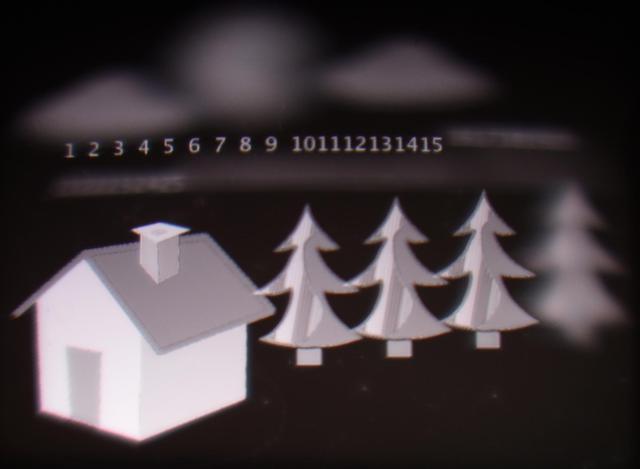}
		\end{subfigure}	
		\hspace{0.3mm}
		\begin{subfigure}[t]{0.24\linewidth}
			\centering
			\includegraphics[height=\figurehieght]{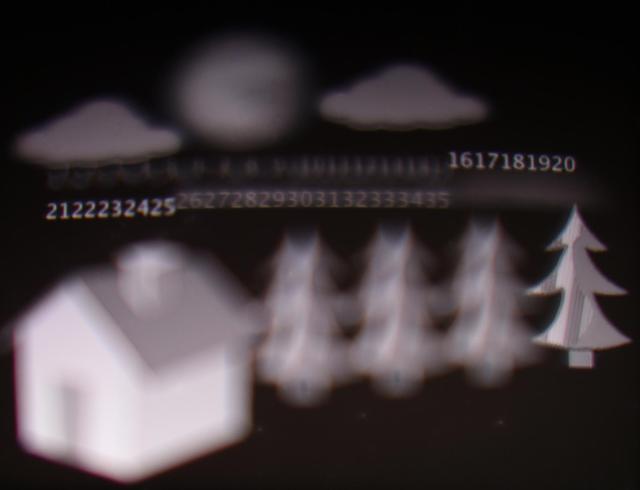}
		\end{subfigure}	
		\begin{subfigure}[t]{0.24\linewidth}
			\centering
			\includegraphics[height=\figurehieght]{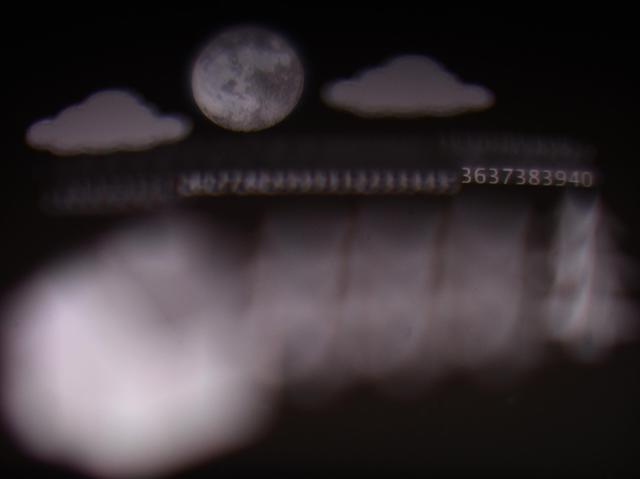}
		\end{subfigure}	
		\vspace{-0.5mm}
		\caption{multifocal display with four focal planes and no depth filtering}
	\end{subfigure}
	\\	
	\def\dataset{4f}
	\begin{subfigure}[t]{\linewidth}
		\centering
		\begin{subfigure}[t]{0.24\linewidth}
			\centering
			\includegraphics[height=\figurehieght]{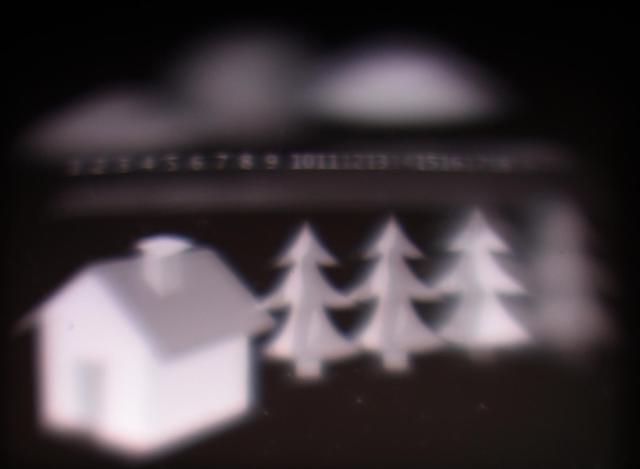}
		\end{subfigure}	
		\hspace{0.5mm}
		\begin{subfigure}[t]{0.24\linewidth}
			\centering
			\includegraphics[height=\figurehieght]{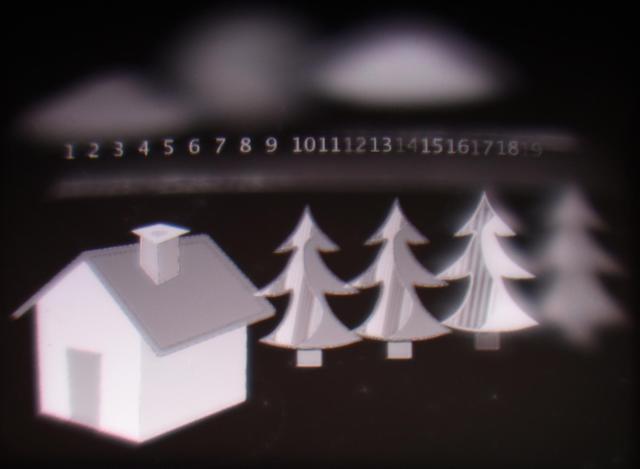}
		\end{subfigure}	
		\hspace{0.3mm}
		\begin{subfigure}[t]{0.24\linewidth}
			\centering
			\includegraphics[height=\figurehieght]{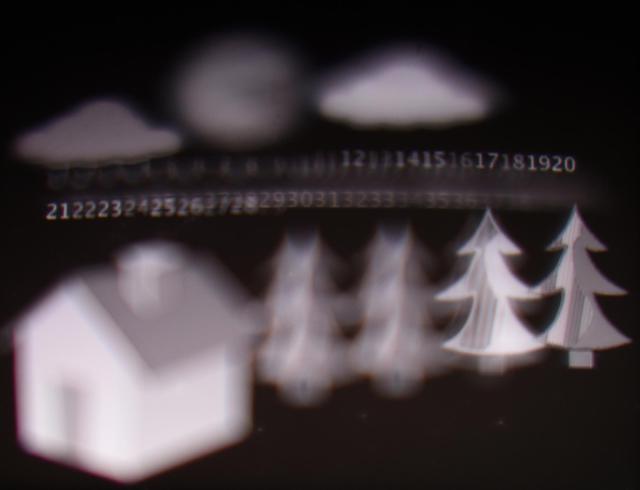}
		\end{subfigure}	
		\begin{subfigure}[t]{0.24\linewidth}
			\centering
			\includegraphics[height=\figurehieght]{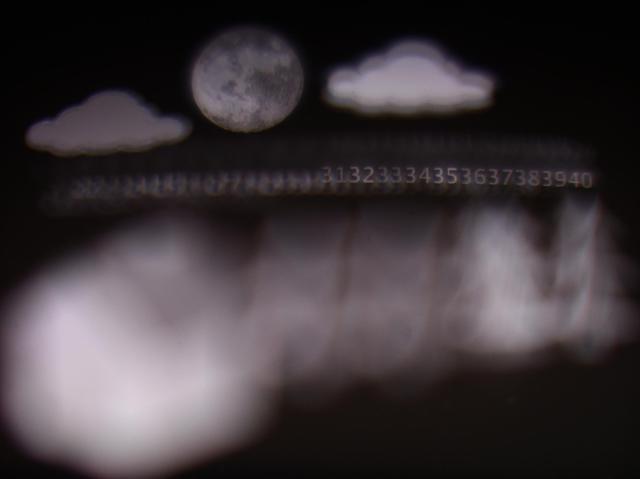}
		\end{subfigure}	
		\vspace{-0.5mm}
		\caption{multifocal display with four focal planes and linear depth filtering}
	\end{subfigure}
	\\
	\def\dataset{40d}
	\begin{subfigure}[t]{\linewidth}
		\centering
		\begin{subfigure}[t]{0.24\linewidth}
			\centering
			\includegraphics[height=\figurehieght]{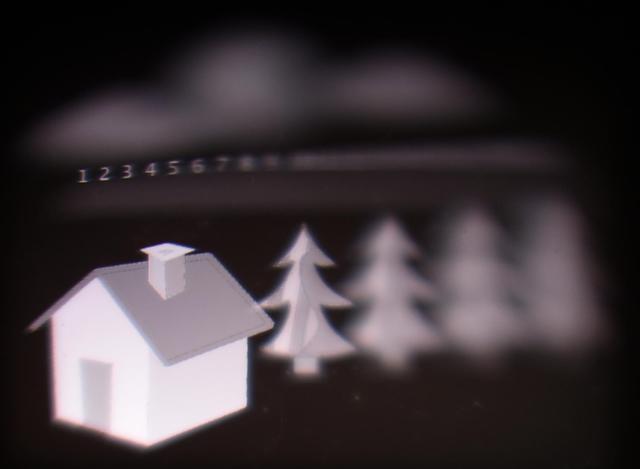}
		\end{subfigure}	
		\hspace{0.5mm}
		\begin{subfigure}[t]{0.24\linewidth}
			\centering
			\includegraphics[height=\figurehieght]{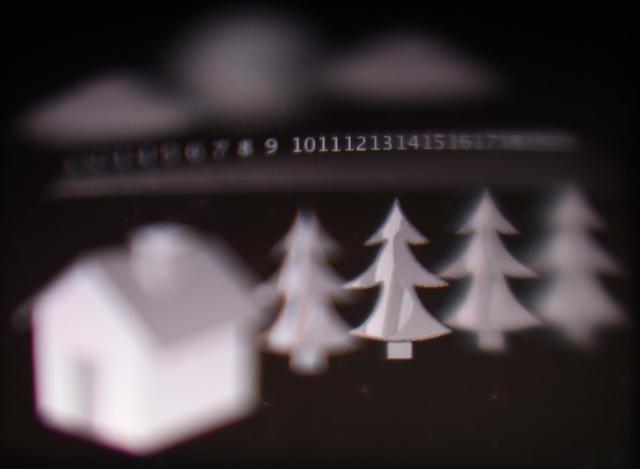}
		\end{subfigure}	
		\hspace{0.5mm}
		\begin{subfigure}[t]{0.24\linewidth}
			\centering
			\includegraphics[height=\figurehieght]{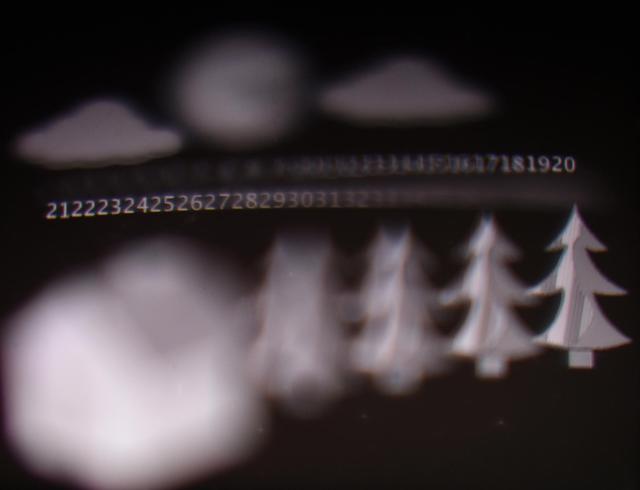}
		\end{subfigure}	
		\begin{subfigure}[t]{0.24\linewidth}
			\centering
			\includegraphics[height=\figurehieght]{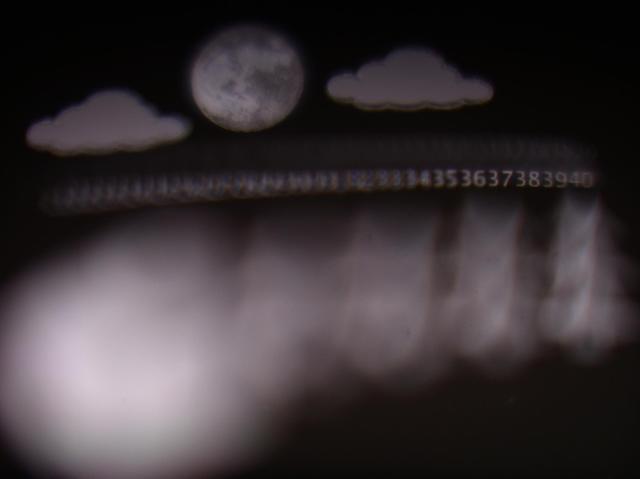}
		\end{subfigure}	
		\vspace{-0.5mm}
		\caption{proposed display with $40$ focal planes and no depth filtering}
	\end{subfigure}
	\vspace{-3mm}
	\caption{Comparison of a typical multifocal display with $4$ focal planes and the proposed display with $40$ focal planes.  The four focal planes of the multifocal display correspond to the $10$th, $20$th, $30$th, and $40$th focal plane.  Images are captured with a $50$ \mm $f/1.4$ lens.  Except for the first column, these focal planes are selected such that the $4$-plane multifocal display (a) is in sharp focus. In the scene, the digits are at their indicated focal planes; the house is at the first focal plane; the trees from left to right are at $5, 10, 15, 20$th focal planes; the clouds and the moon are at $30, 35, 40$th, respectively. 
	}
	\label{figure: bigfont}
\end{figure*}

\subsection{Benefits of Dense Focal Stacks}
To evaluate the benefit provided by dense focal stacks, we simulate two multifocal displays, one with 4 focal planes and the other with 40 focal planes.
The 40 focal planes are distributed uniformly in diopter from 0 to 4 diopters, and the 4-plane display has focal planes at the depth of the 5th, 15th, 25th, and 35th focal planes of the 40-plane display.
The scene is composed of 28 resolution charts, each at a different depth from 0 to 4 diopters (please refer to the supplemental material for figures of the entire scene).
The dimension of the scene is $1500\times2000$ pixels.

We render the scene with three methods:  
\begin{itemize}[leftmargin=*]
	\item \textit{No depth filtering}:  We directly quantize the depth channel of the images to obtain the focal planes of different depths.  
	\item \textit{Linear depth filtering}:  Following~\citep{akeley2004stereo}, we apply a triangular filter on the focal planes based on their depths. 
	\item \textit{Optimization-based filtering}:  We follow the formulation proposed in~\cite{mercier2017multifocal}.  We first rendered normally the desired retinal images focused at 81 depths uniformly distributed across 0 to 4 diopters in the scene with a pupil diameter of 4 mm.  
	Then we solve the optimization problem to get the content to be displayed on the focal planes. 
	We initialize the optimization process with the results of direct quantization and perform gradient descent with $500$  iterations to ensure convergence.
\end{itemize}

The perceived images of the resolution chart at $0.02$ diopters are shown in Figure~\ref{figure: res results}; a plane at $0.02$ diopters is on a focal plane of the 40-plane display and is at the furthest inter-focal plane of the 4-plane display.
Note that we simulate the results with pupil diameter of $4$ \mm, which is a typical value used to simulated retinal images of human eyes.

As can be seen from the results, the perceived images of the 40-plane display closely follow those of the ground truth --- with high spatial resolution if the camera is focused on the plane (Figure~\ref{figure: res results}a) and natural retinal blur when the camera is not focused (Figure~\ref{figure: res results}b).
In comparison, at its inter-plane location (Figure~\ref{figure: res results}a), the 4-plane display has much lower spatial resolution than the other display, regardless of the depth filtering methods applied.
These results verify our analysis in Section~\ref{sec: how many}.

To evaluate the benefit provided by dense focal stacks in providing higher spatial resolution when the eye is focused at an inter-plane location, we implement four multifocal displays with 4, 20, 30 and 40 focal planes, respectively, on our prototype.
The 4-plane display has its focal planes on the $5, 15, 25, 35$th focal planes of the 40-plane display, and the 20-plane display has its focal planes on all the odd-numbered focal planes.
We display a resolution chart on the fifth focal plane of the 40-plane display; this corresponds to a depth plane that all three displays can render.

To compare the worst-case scenario where an eye focuses on an inter-plane location, we focus the camera at the middle of two consecutive focal planes of each of the displays.
In essence, we are reproducing the effect of VAC where the vergence cue forces the ocular lens to focus on an inter-focal plane. For the 40-plane display, this is between focal planes five and six. For the 20-plane display, this is on the sixth focal plane of the 40-plane display. And for the 4-plane display, this is on the tenth focal plane of the 40-plane display. 
We also focus the camera on the estimated inter-plane location of a 30-plane display.
The results captured by a camera with a $50$ \mm $f/1.4$ lens are shown in Figure~\ref{figure: interplane}.
As can be seen, the higher number of focal planes (smaller focal-plane separation) results in higher spatial resolution at inter-plane locations.

Next, we compare our prototype with a 4-plane multifocal display on a real scene.
Note that we implement the 4-plane multifocal display with our 40-plane prototype by showing contents on the $10, 20, 30, 40$th focal planes.
The  images captured by the camera are shown in Figure~\ref{figure: bigfont}.  
For the 4-plane multifocal display, when used without linear depth filtering, virtual objects at multiple depths are focus/defocus as groups; when used with linear depth filtering, same objects appearing in two focal planes reduces the visibility and thereby lowers the resolution of the display.
In comparison, the proposed method produces smooth focus/defocus cues across the range of depths, and the perceived images at inter-plane locations (\eg $0.25$ m) have  higher spatial resolution than the 4-plane display.

\begin{figure*}[t]
	\centering
	\begin{subfigure}[t]{0.245\linewidth}
		\centering
		\includegraphics[width=\linewidth]{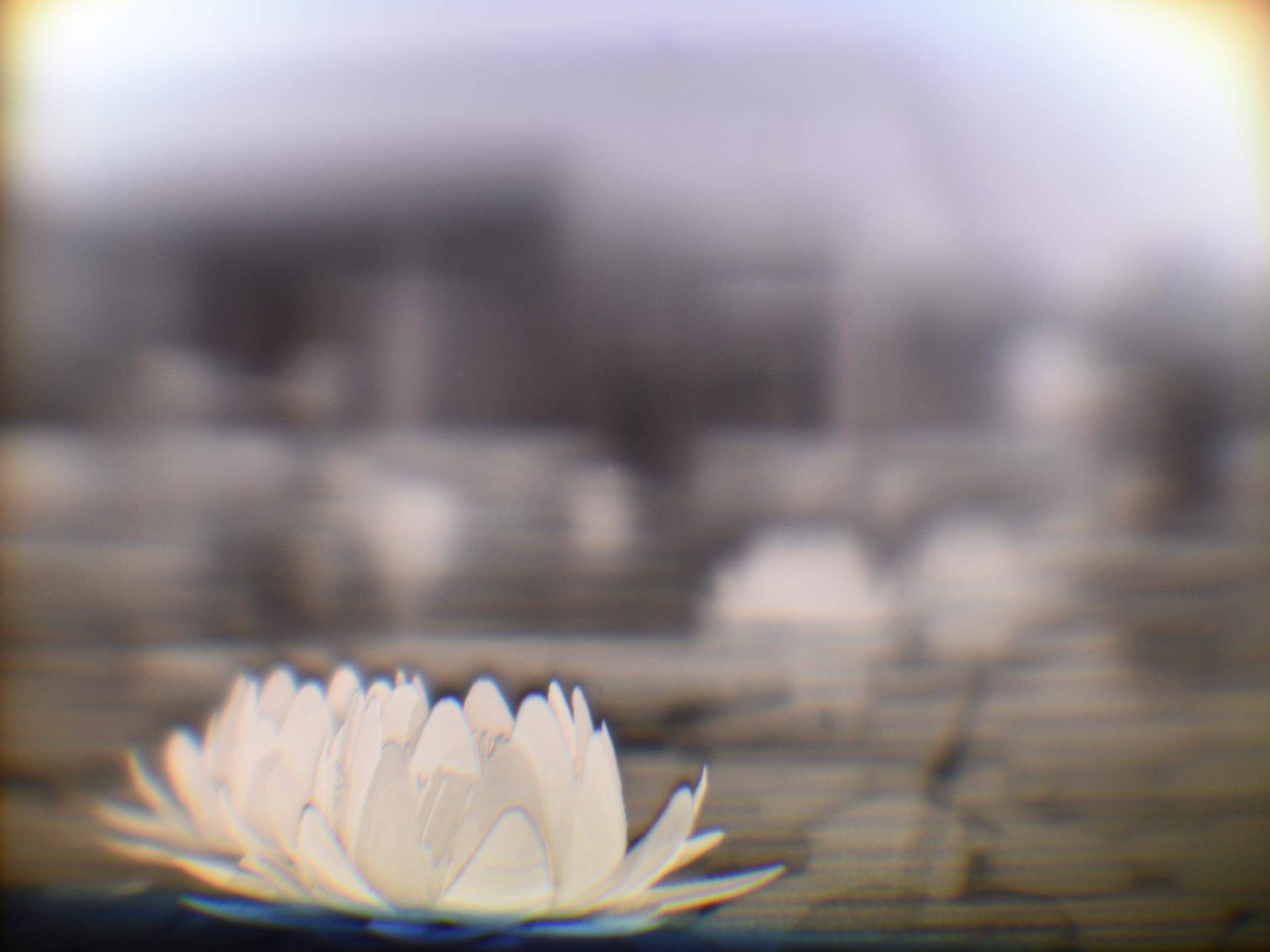}
	\end{subfigure}
	\begin{subfigure}[t]{0.245\linewidth}
		\centering
		\includegraphics[width=\linewidth]{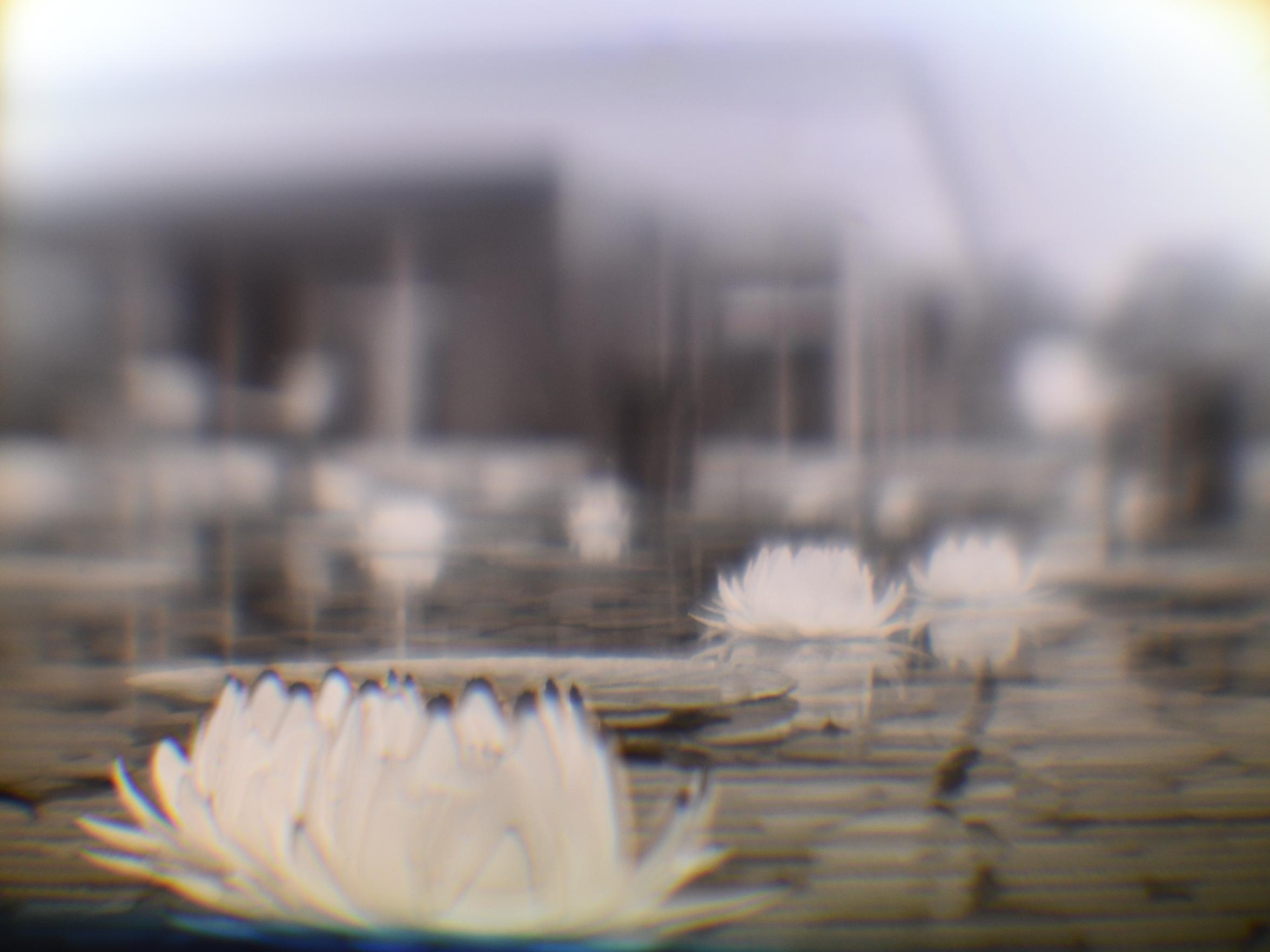}
	\end{subfigure}
	\begin{subfigure}[t]{0.245\linewidth}
		\centering
		\includegraphics[width=\linewidth]{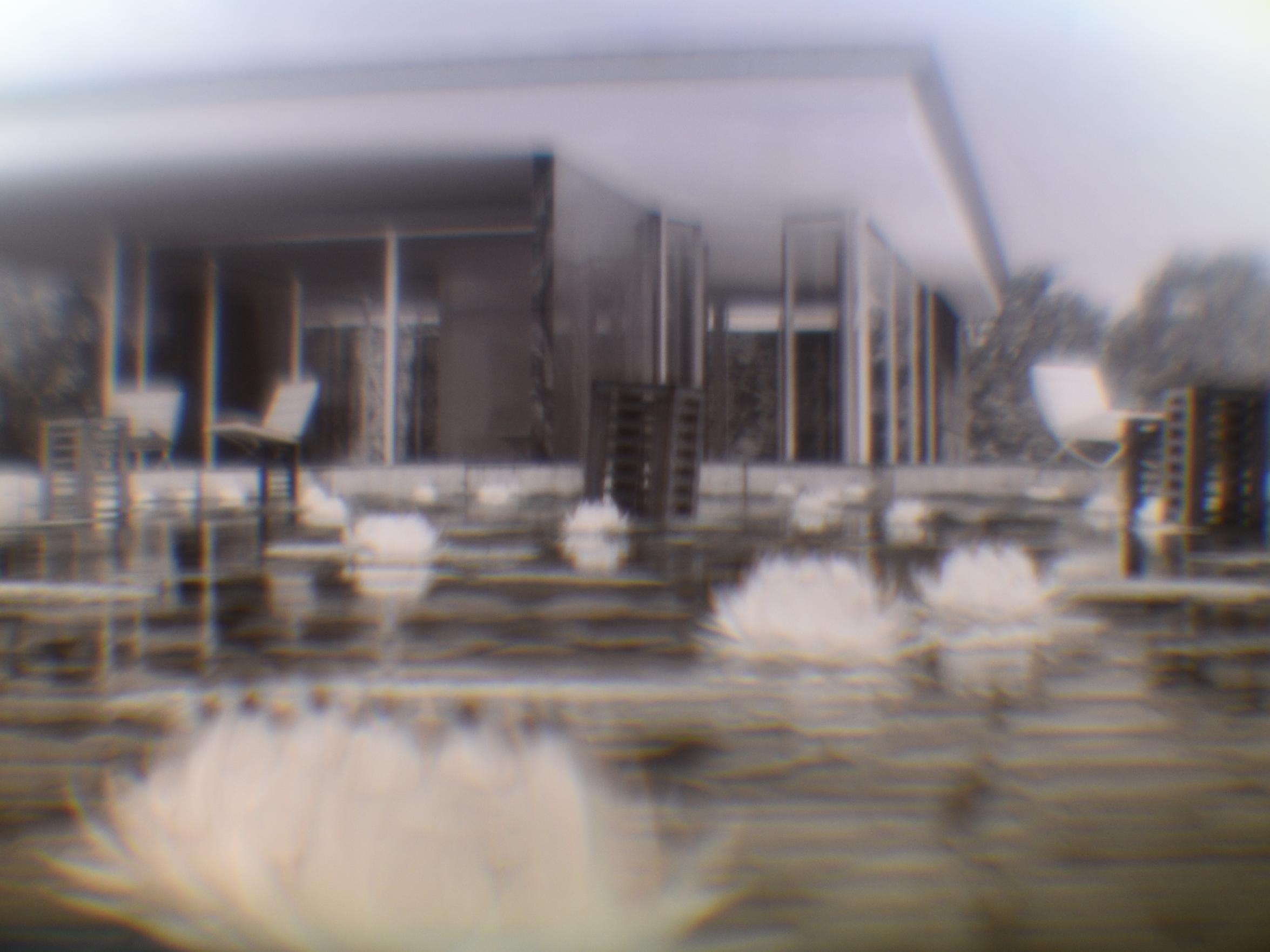}
	\end{subfigure}
	\begin{subfigure}[t]{0.245\linewidth}
		\centering
		\includegraphics[width=\linewidth]{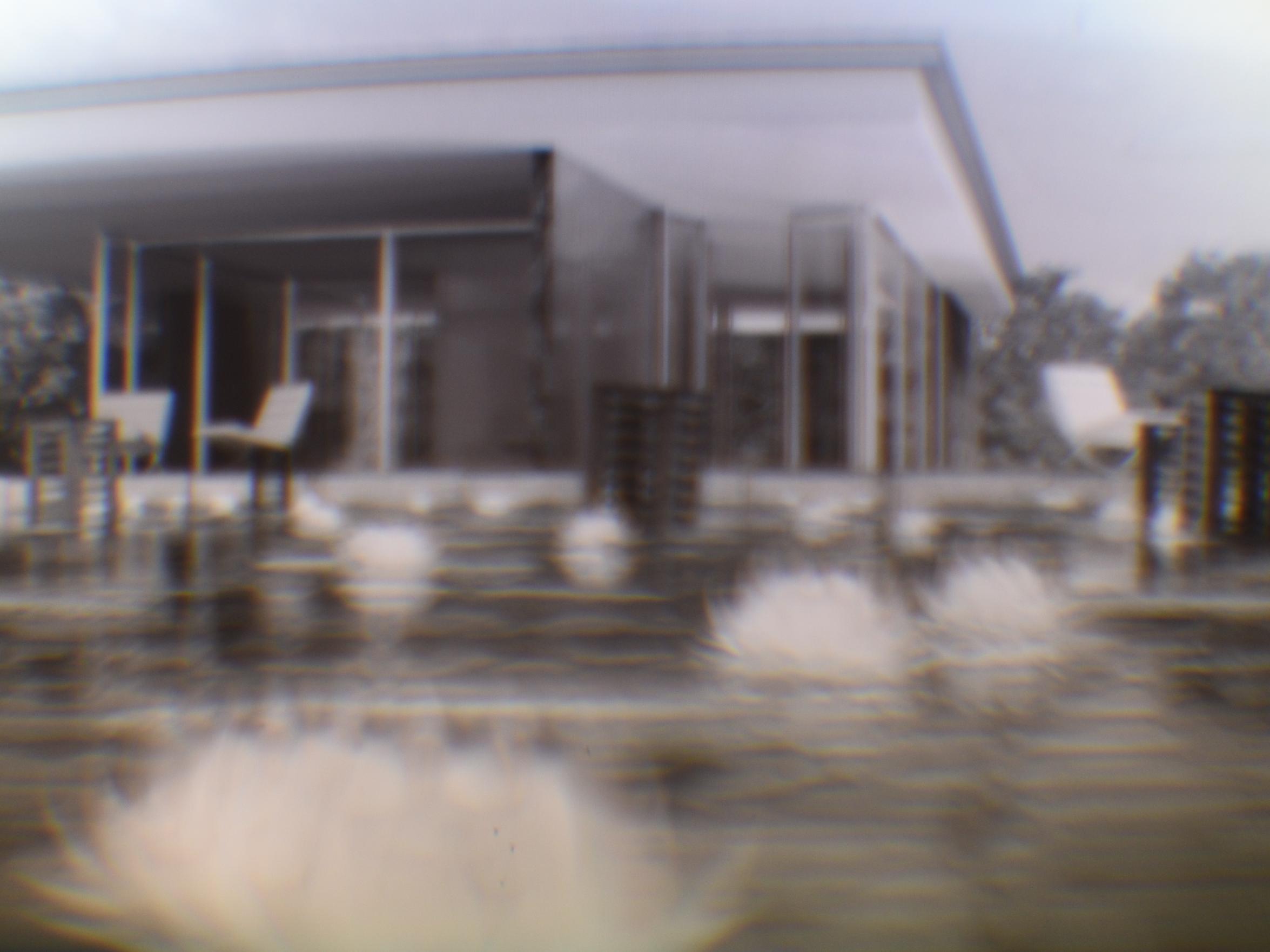}
	\end{subfigure}
	\vspace{-3mm}
	\caption{Captured images focusing from near (shown at left) to far (shown at right) of a simulated scene rendered by Blender.  The scene depth ranges from $50$ \cm (the flower at the bottom left) to infinity (the sky).  The camera has a $50$ \mm $f/1.4$ lens. Three-dimensional scene courtesy eMirage.}
	\label{figure: pabellon}
\end{figure*}

Finally, we render a more complex scene~\citeA{pabellon} using Blender.
From the rendered all-in-focus image and its depth map, we perform linear filtering and display the results with the prototype.
Focus stack images captured using a camera are shown in Figure~\ref{figure: pabellon}.
We observe realistic focus and defocus cues in the captured images.

\section{Conclusion}
This paper provides a simple but effective technique for displaying virtual scenes that are made of a dense collection of focal planes.
Despite the bulk of our current prototype, the proposed tracking technique is fairly straightforward and extremely amenable to miniaturization.
We believe that the system proposed in the paper for high-speed tracking could spur innovation in not just virtual and augmented reality systems but also in traditional light field displays.

\begin{acks}
	The authors acknowledge support via the NSF CAREER grant CCF-1652569 and a gift from Adobe Research.
\end{acks}

\bibliographystyle{ACM-Reference-Format}
\bibliography{main}

\appendix

\section{Light Field Analysis}
\label{appendix: proof}

This section provides a detailed derviation of the analysis discussed in Section~\ref{sec: how many} of the main paper in detail.
This analysis follows closely to the one in~\citep{narain2015optimal}.
A notable difference however is that we provide analytical expressions for the perceived spatial resolution (Equation~\eqref{eq: spatial res} in the main paper) and the minimum number of focal planes required (Equation~\eqref{eq: num planes}), whereas they only provide numerical results.
For simplicity, we consider a flatland where a light field is two-dimensional and is parameterized by intercepts with two parallel axes, $x$ and $u$.  
The two axes are separated by $1$ unit, and for each $x$, we align the origin of $u$-axis to $x$.  
We model the human eye with a camera model that is composed of a finite-aperture lens and a sensor plane $d_e$ away from the lens, as that used by \citeN{mercier2017multifocal} and \citeN{sun2017perceptually}
We assume that the display and the sensor emits and receives light isotropically so that each pixel on the display uniformly emits light rays toward every direction, and vice versa for the sensor.

\begin{figure*}[!thh]
	\centering
	\includegraphics[width=\linewidth]{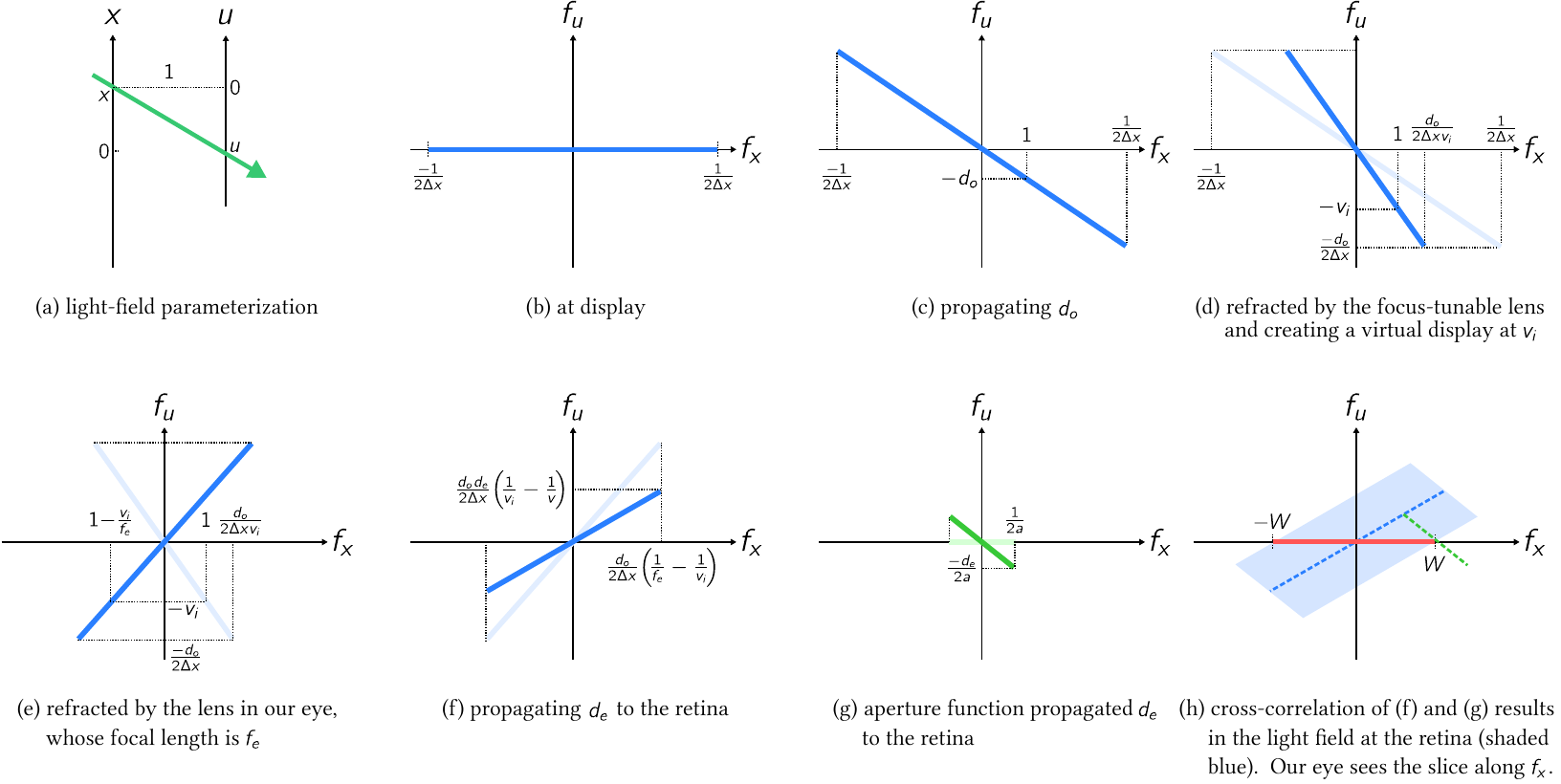}
	\caption{Fourier transform of the 2-dimensional light field at each stage of a multifocal display.  The display is assumed to be isotropic and has pixels of pitch $\Delta x$.  (a) Each light ray in the light field  is characterized by its intercepts with two parallel axes, $x$ and $u$, which are separated by $1$ unit, and the origin of the $u$-axis is relative to each individual value of $x$. (b) With no angular resolution, the light field spectrum emitted by the display is a flat line on $f_x$.  We focus only on the central part ($|f_x| \le \frac{1}{2\Delta x}$).  (c) The light field propagates $d_o$ to the tunable lens, causing the spectrum to shear along $f_u$.  (d) Refraction due to the lens corresponds to shearing along $f_x$, forming a line segment of slope $-v_i$, where $v_i$ is the depth of the focal plane.  (e,f) Refraction by the lens in our eye  and propagation $d_e$ to the retina without considering the finite aperture of the pupil.  (g) The spectrum of the pupil function propagates $d_e$ to the retina.  (h) The light field spectrum on the retina with a finite aperture is the 2-dimensional cross-correlation between (f) and (g).  According to Fourier slice theorem, the spectrum of the perceived image is the slice along $f_x$, shown as the red line.  The diameter of the pupil and the slope of (f), which is determined by the focus of the eye and the virtual depth $v_i$, determine the spatial bandwidth, $W$, of the perceived image.  }
	\label{figure: light field supp}
\end{figure*}

\paragraph{Light Field Generated by a Display} Let us decompose the optical path from the display to the retina (sensor) and examine the effect in frequency domain due to each component.
Due to the finite pixel pitch, the light field creates by the display can be model as 
\[
\ell_d(x,u) =  \left( \rect\left(\frac{x}{\Delta x}\right) \ast \ell_t(x,u=0) \right) \times \sum_{m=-\infty}^{\infty} \delta(x - m \Delta x), 
\]
where $\ast$ represents two-dimensional convolution, $\Delta x$ is the pitch of the display pixel, and $\ell_t$ is the target light field.
The Fourier transform of $\ell_d(x,u)$ is 
\[
L_d(f_x, f_u) = \left( \sinc(\Delta x f_x)  \ \delta(f_u) \  L_t(f_x, f_u) \right) * \sum_{m=-\infty}^{\infty} \delta(f_x - \frac{m}{\Delta x}).
\]
The finite pixel pitch acts as an anti-aliasing filter and thus we consider only the central spectrum replica ($m=0$).
Also, we assume $|L_t(f_x, f_u)| = 0$ for all $|f_x| \ge \frac{1}{2 \Delta x}$ to avoid aliasing.
Since the light field is nonnegative, or $\ell_d \ge 0$, we have $\left| L_t(f_x, f_u) \right| \le L_t(0, 0)$.  %
Therefore, we have 
\begin{align}
&	\left| L_d(f_x, f_u) \right|  \le L_t(0, 0) \ |\sinc(\Delta x f_x)|  \ \delta(f_u),  &  |f_x| \le \frac{1}{2 \Delta x} \\
&	\left| L_d(f_x, f_u) \right|  = 0, & \mbox{otherwise}.
\end{align}	
Therefore,  in the ensuing derivation, we will focus on the upper-bound 
\[ \widehat{L}_d = \sinc(\Delta x f_x) \, \delta(f_u) \, \rect\left(  \frac{f_x}{\Delta x} \right). \]
The light field spectrum $\widehat{L}_d$ forms a line segment parallel to $f_x$, as plotted in Figure~\ref{figure: light field supp}a.

\paragraph{Propagation to the eye}
After leaving the display, the light field propagates $d_o$ and get refracted by the focus-tunable lens before reaching the eye.  
Under first-order optics, there operations can be modeled by coordinate transformation of the light fields~\cite{hecht2002optics}. 
Let $\x = \left[ x \ u\right]^\top$. 
After propagating a distance $d_o$,  the output light field is a reparameterization of the input light field  and can be represented as  
\[
\ell_o(\x) = \ell_i\left(P_{d_o}^{-1}\x\right), \mbox{where } P_{d_o} = 
\begin{bmatrix}
1 & d_o \\ 0 & 1
\end{bmatrix}.
\]
After refracted by a thin lens with focal length $f$,  the output light field right after the lens is
\[
\ell_o(\x) = \ell_i\left(R_{f}^{-1}\x\right), \mbox{where } R_f = 
\begin{bmatrix}
1 & 0 \\ \frac{-1}{f} & 1
\end{bmatrix}.
\]
Since $P_{d_o}$ and $R_f$ are invertible, we can use the stretch theorem of $d$-dimensional Fourier transform to analyze their effect in the frequency domain.
The general stretch theorem states that: 
Let $\x \in \R^d$, $\mF(\cdot)$ be the Fourier transform operator, and $A \in \R^{d \times d}$ be any invertible matrix.  We have
\[
\mF \left( \ell(A \x) \right) = \frac{1}{\left|\mbox{det } A\right|}\, L(A^{-\top} \f),
\]
where $L$ is the Fourier transform of $\ell$, $\f \in \R^d$ is the variable in frequency domain, $\mbox{det }A$ represents determinant of $A$, and $A^{-\top} = \left({A^\top}\right)^{-1} = \left({A^{-1}}\right)^\top$.
By applying the stretch theorem to $P_{d_o}$ and $R_f$, we can see that propagation and refraction shears the Fourier transform of the light field along $f_u$ and $f_x$, respectively, as shown in Figure~\ref{figure: light field supp}c-d.

\paragraph{Light Field Incident on the Retina} After reaching the eye, the light field $\ell_o$ is partially blocked by the pupil, refracted by the lens of the eye,  propagates $d_e$ to the retina, and finally integrated through all directions to form an image.
The light field reaching the retina can be represented as 
\[
\ell_e(\x) = \ell_a\left( R_{f_e}^{-1} P_{d_{e}}^{-1} \x \right), \mbox{where } \ell_a(\x) = \rect\left(\frac{x}{a}\right) \ell_o(\x), 
\]
and $a$ is the diameter of the pupil.  
To understand the effect of the aperture, we analyze a more general situation where the light field is multiplied with a general function $h(\x)$ and transformed by an invertible $T$ with unit determinant.
By multiplication theorem, we have 
\[
\ell_a(\x) = h(\x) \times \ell_o(\x)  \  \xleftrightarrow{\mF} \  L_a(\f) = H(\f) * L_o(\f).
\]
Thereby, 
\begin{align}
L_a(T \f) & = \int L_o(\p) H(\p - T\f) \dd{\p} = \int L_o(\p) H\left(T \left(T^{-1}\p - \f \right) \right) \dd{\p} \nonumber \\
  & =  \int L_o\left( T(\q + \f) \right) H\left(T \q \right) \left| \pdv{\p}{\q} \right| \dd{\q} = L_o^{(T)} \otimes H^{(T)}(\f) , \label{eq: cross correlation} 
\end{align}
where we use a change of variable by setting $\q = T^{-1}\p -\f$, and the last equation holds because $\left| \pdv{\p}{\q} \right|  = \mbox{det }T = 1$.
Equation~\eqref{eq: cross correlation} relates the effect of the aperture directly to the output light field at the retina:  The spectrum of the output light field is the cross correlation between the transformed (refracted and propagated) input spectrum with full aperture and the transformed spectrum of the aperture function.
The result is important since it significantly simplifies our analysis, and as a result, we are able to derive an analytical expression of spatial resolution and number of focal planes needed.

In our scenario, we have $T = \left(R_{f_e}^{-1} P_{d_e}^{-1} \right)^{-\top}$.  
For a virtual display at $v_i$, $\ell_o(\x)$ is a line segment of slope $-v_i$ within $x \in [\frac{-1}{2 \Delta x_i}, \frac{1}{2 \Delta x_i}]$, where $\Delta x_i = \left| \frac{v_i}{d} \right| \Delta x$ is the magnified pixel pitch.
According to Equation~\eqref{eq: cross correlation}, $L_e(\f) = L_a(T\f)$ is simply the cross correlation of  $L_o(T \f)$ and $\sinc(T \f)$.
After transformation, $L_a(T\f)$ is a line segment of slope $\frac{d_e v_i - (d_e +v_i)f_e}{ v_i - f_e}$, where $|x| \le \left| \left(\frac{v_i}{f_e} - 1\right)\frac{1}{\Delta x_i} \right| $.
Similarly, $\sinc(T \f)$ is a line segment with slope $-d_e$ within $|x| \le \frac{1}{2a}$.
Note that we only consider $|x| \le \frac{1}{2a}$ because the cross-correlation result at the boundary has value $\sinc(0.5) \times \sinc(0.5) \approx 0.4$.  
Since $\sinc(x)$ function is monotonically decreasing for $|x| \le 1$, the half-maximum spectral bandwidth ($|L_e(\f)| = 0.5$) must be  within the region.
Let the depth the eye is focusing at be $v$.  We have $\frac{1}{v} + \frac{1}{d_e} = \frac{1}{f_e}$.  
When $v = v_i$, we can see from the above expression that $L_a(T \f)$ is a flat segment within $|f_x| \le \frac{1}{2 M \Delta x}$, where $M = \frac{d_e}{d_o}$ is the overall magnification caused by the focus-tunable lens and the lens of the eye.  
From Fourier slice theorem, we know that the spectrum of the image is simply the slice $L_a(T \f)$ along $f_x$. 
In this case, the aperture has no effect to the final image, since the cross correlation does not extend or reduce the spectrum along $f_x$, and the final image has the highest spatial resolution $\frac{1}{2 M \Delta x}$.

Suppose the eye does not focus on the virtual display, or $v \neq v_i$.  
In the case of a full aperture ($a \rightarrow \infty$), the resulted image will be a constant DC term (completely blurred) because the slice along $f_x$ is a delta function at $f_x = 0$.
In the case of finite aperture diameter $a$, with a simple geometric derivation (see Figure~\ref{figure: light field supp}h), we can show by simple geometry that the bandwidth of the $f_x$-slice of $L_e(\f)$, or equivalently, the region $\left\{ f_x | L_e(f_x, 0) \ge 0.5 \right\}$, is bounded by $|f_x| \le W$.
And we have
\begin{equation}
W = 
\begin{cases}
\frac{d_o}{2 \Delta x d_e}, & \mbox{if} \left| \frac{1}{v_i} - \frac{1}{v}  \right| \le \frac{\Delta x}{a d_o} \\
\frac{1}{2 a d_e} \left| \frac{1}{v} - \frac{1}{v_i} \right|^{-1}, & \mbox{otherwise}.
\end{cases}
\label{eq: appendix W}
\end{equation}
Thereby, based on Fourier slice theorem, the bandwidth of the retinal images is bounded by $W$.
\section{Other Discussions}

\subsubsection*{Color} Color display can be implemented by using a three color LED and cycling through them using time division multiplexing.
This would lead to loss in time-resolution or focal stack resolution by a factor of $3$.
This  loss in resolution can be completely avoided with OLED-based high speed displays since each group of pixels automatically generate the desired image at each focal stack.

\begin{figure}[t]
	\centering
	\begin{subfigure}[t]{\linewidth}
		\centering
		\includegraphics[width=\linewidth]{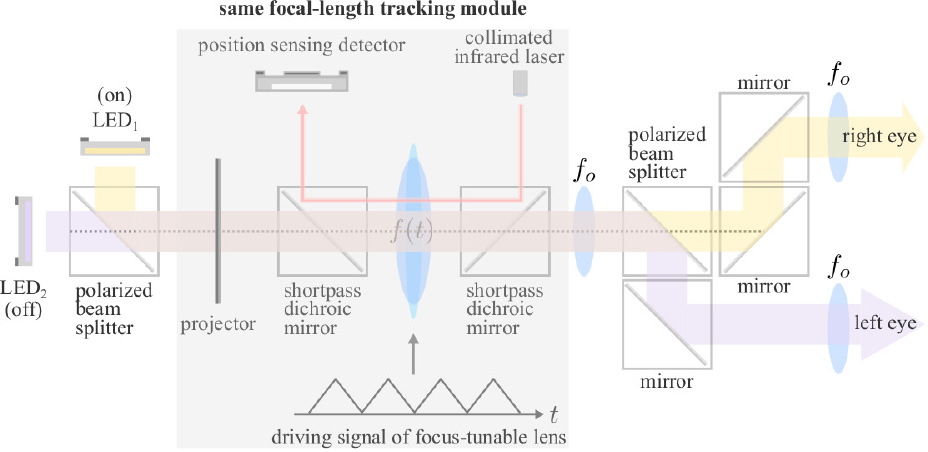}
		\vspace{-5mm}
		\caption{\footnotesize Schematic}
	\end{subfigure}	
	\\
	\vspace{3mm}
	\begin{subfigure}[t]{\linewidth}
		\centering
		\includegraphics[width=\linewidth]{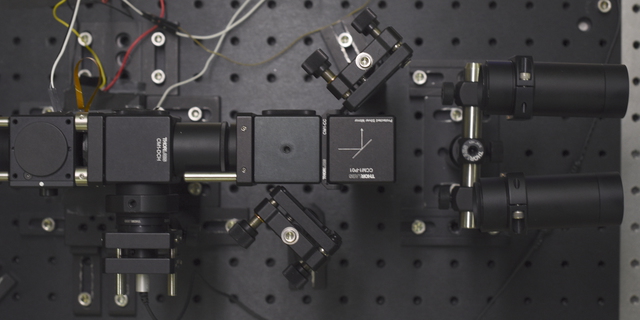}
		\vspace{-5mm}
		\caption{\footnotesize The front portion of the stereo version of the prototype}
	\end{subfigure}	
	\vspace{-1mm}
	\caption{Supporting stereo with a single focus-tunable lens and focus-length tracking module.  The design utilizes two LEDs as light sources of the DMD projector.  Two polarized beam splitters are used to create dedicated light path for LED$_1$ (to the right eye) and LED$_2$ (to the left eye).   To show the content on the DMD to the right eye, only LED$_1$ is turned on, and vice versa.  To account for the extra distance created by the optics, we use two 4f systems (sharing the first lens) with $f = 75$ \mm to bring both eyes virtually to the aperture of the focus-tunable lens.  }
	\label{fig: stereo}
\end{figure}

\subsubsection*{Stereo virtual display}

The proposed method can be extended to support stereo virtual reality displays.
The most straight-forward method is to use two sets of the prototypes, one for each eye.  
Since all focal planes are shown in each frame, there is no need to synchronize the two focus-tunable lenses.
It is also possible to create a stereo display with a single focus tunable lens and a single tracking module;
the design for this is shown in Figure~\ref{fig: stereo}.
This design trades half of the focal planes to support stereo, and  thereby, only requires one set of the prototype and additional optics.
Polarization is used to ensure that each eye only sees the scene that is meant to see.

\def \one {gt}
\def \onedes {ground truth}
\def \two {direct4}
\def \twodes {4-plane, direct quantization}
\def \three {linear4}
\def \threedes {4-plane, linear filtering}
\def \four {inset1}
\def \fourdes {inset: $D=0.02$}
\def \five {opt4}
\def \fivedes {4-plane, optimization-based filtering}
\def \six {direct40}
\def \sixdes {40-plane, direct quantization}
\def \seven {opt40}
\def \sevendes {40-plane, optimization-based filtering}
\def \eightdes {modulation transfer function of $D=0.02$}

\begin{figure*}[t]
	\centering
	\def \folder {e_1}
	\begin{subfigure}[t]{\linewidth}
		\centering
		\begin{subfigure}[t]{0.32\linewidth}
			\centering
			\includegraphics[width=\linewidth]{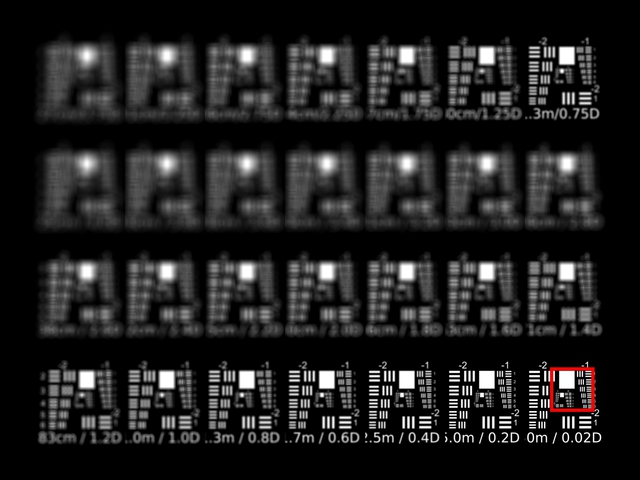}
			\vspace{-6mm}
			\caption*{\footnotesize \onedes}
		\end{subfigure}	
		\begin{subfigure}[t]{0.32\linewidth}
			\centering
			\includegraphics[width=\linewidth]{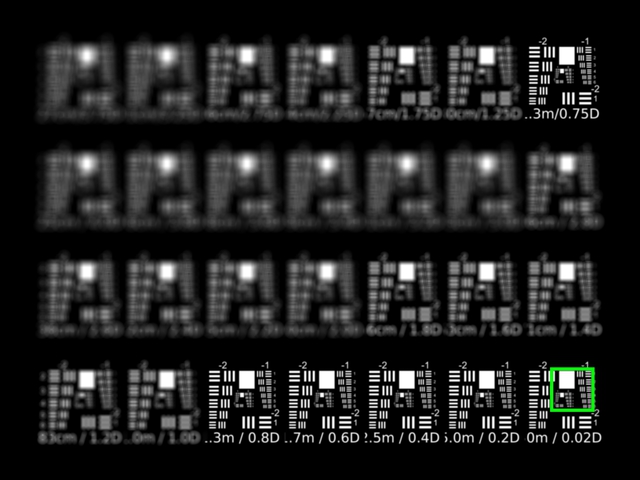}
			\vspace{-6mm}
			\caption*{\footnotesize \twodes}
		\end{subfigure}	
		\begin{subfigure}[t]{0.32\linewidth}
			\centering
			\includegraphics[width=\linewidth]{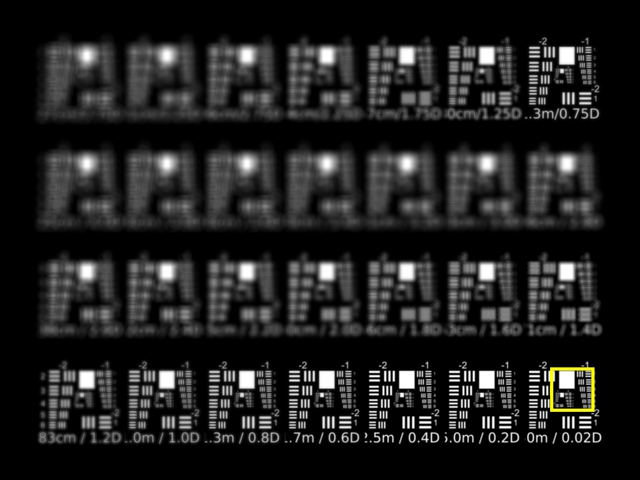}
			\vspace{-6mm}
			\caption*{\footnotesize \threedes}
		\end{subfigure}	
		\\
		\vspace{1mm}
		\begin{subfigure}[t]{0.32\linewidth}
			\centering
			\includegraphics[width=\linewidth]{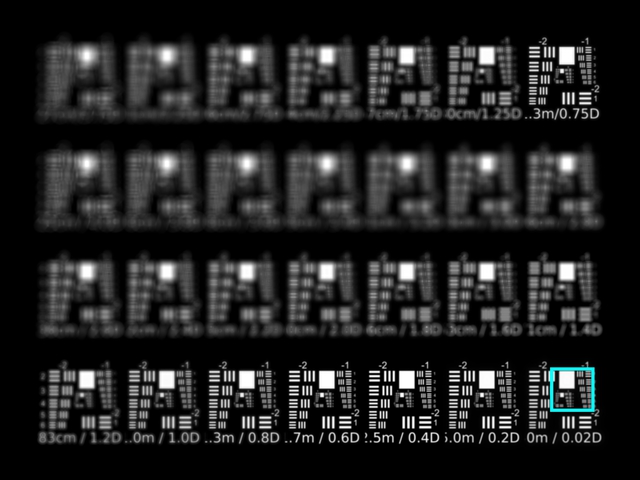}
			\vspace{-6mm}
			\caption*{\footnotesize \fivedes}
		\end{subfigure}	
		\begin{subfigure}[t]{0.32\linewidth}
			\centering
			\includegraphics[width=\linewidth]{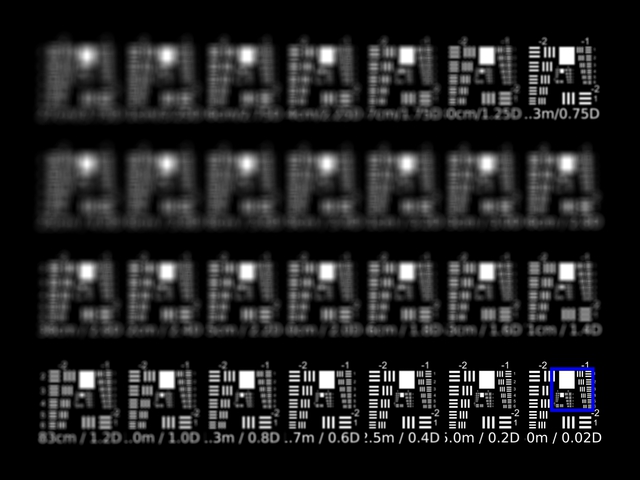}
			\vspace{-6mm}
			\caption*{\footnotesize \sixdes}
		\end{subfigure}	
		\begin{subfigure}[t]{0.32\linewidth}
			\centering
			\includegraphics[width=\linewidth]{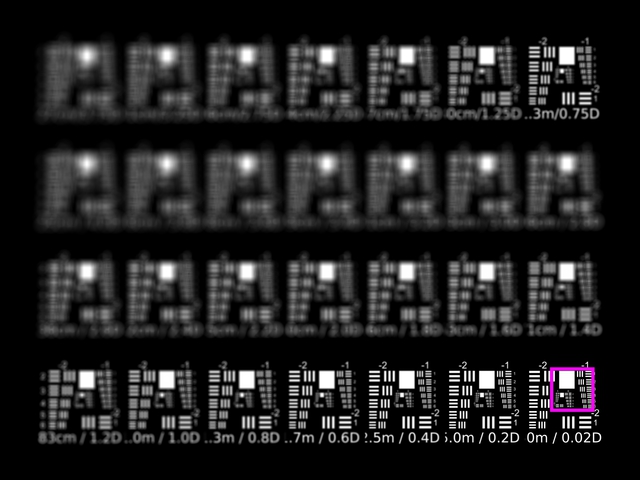}
			\vspace{-6mm}
			\caption*{\footnotesize \sevendes}
		\end{subfigure}	
		\vspace{-1mm}
		\caption{\footnotesize Camera focuses at $0.02$ diopters (off-plane location of the 4-plane display)}
	\end{subfigure}	
	\\
	\vspace{2mm}
	\def \folder {e_19}
	\begin{subfigure}[t]{\linewidth}
		\centering
		\begin{subfigure}[t]{0.32\linewidth}
			\centering
			\includegraphics[width=\linewidth]{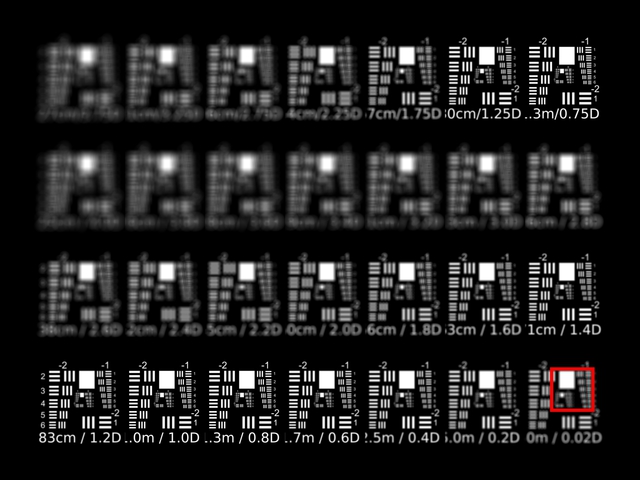}
			\vspace{-6mm}
			\caption*{\footnotesize \onedes}
		\end{subfigure}	
		\begin{subfigure}[t]{0.32\linewidth}
			\centering
			\includegraphics[width=\linewidth]{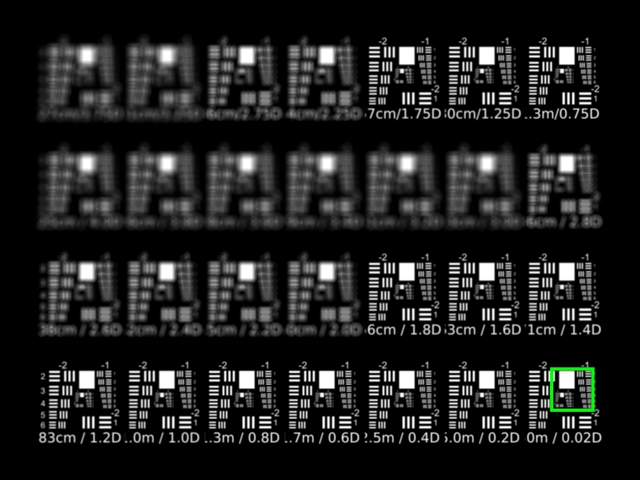}
			\vspace{-6mm}
			\caption*{\footnotesize \twodes}
		\end{subfigure}	
		\begin{subfigure}[t]{0.32\linewidth}
			\centering
			\includegraphics[width=\linewidth]{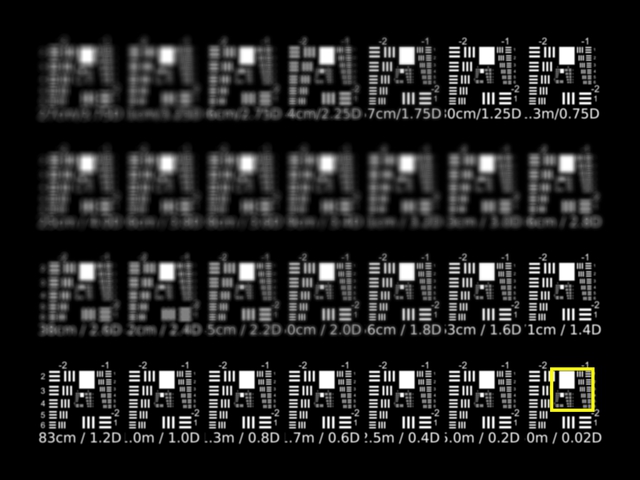}
			\vspace{-6mm}
			\caption*{\footnotesize \threedes}
		\end{subfigure}	
		\\
		\vspace{1mm}
		\begin{subfigure}[t]{0.32\linewidth}
			\centering
			\includegraphics[width=\linewidth]{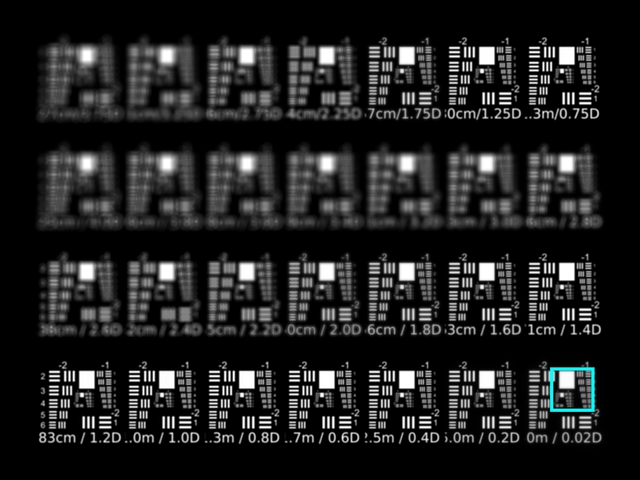}
			\vspace{-6mm}
			\caption*{\footnotesize \fivedes}
		\end{subfigure}	
		\begin{subfigure}[t]{0.32\linewidth}
			\centering
			\includegraphics[width=\linewidth]{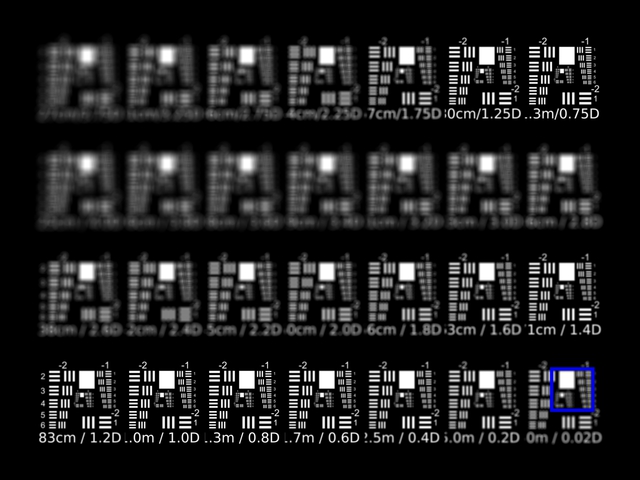}
			\vspace{-6mm}
			\caption*{\footnotesize \sixdes}
		\end{subfigure}	
		\begin{subfigure}[t]{0.32\linewidth}
			\centering
			\includegraphics[width=\linewidth]{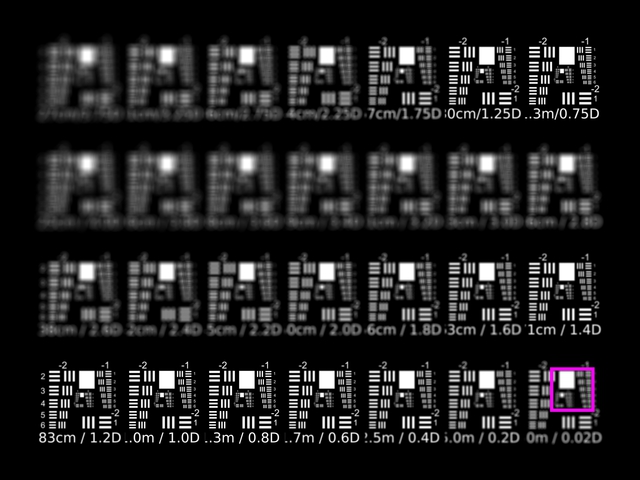}
			\vspace{-6mm}
			\caption*{\footnotesize \sevendes}
		\end{subfigure}	
		\vspace{-1mm}
		\caption{\footnotesize Camera focuses at $0.9$ diopters (off-plane location for both displays)}
	\end{subfigure}	
	\vspace{-2mm}
	\caption{Simulation results of 4-plane and 40-plane multifocal displays with direct quantization, linear depth filtering, and optimization-based filtering.  The indicated regions are used to plot Figure~\ref*{figure: res results} in the paper. 
	}
	\label{figure: res results all}
\end{figure*}	 %

\section{Simulated Scene}
\label{sec: simulated scene}
Figure~\ref{figure: res results all} shows the simulated images of Figure~\ref{figure: res results} in the paper with full field-of-view.
There are 28 resolution charts located at various depths from 0 to 4 diopters (as indicated by beneath each of them). 
In the figure, we plot the ground-truth rendered images and simulated retinal images when focused on 0.02 diopters and 0.9 diopters. 
Rest of the focus stack can be seen in the supplemental video.
 
\end{document}